\documentclass[10pt,twocolumn,letterpaper]{article}

\usepackage{iccv}
\usepackage{times}
\usepackage{epsfig}
\usepackage{graphicx}
\usepackage{amsmath}
\usepackage{amssymb}
\usepackage{tensor}
\usepackage[caption=false]{subfig}
\usepackage{diagbox}
\usepackage{enumitem}
\usepackage{rotating} 
\usepackage{cases}
\usepackage{bm}
\usepackage{color}
\usepackage{colortbl}
\usepackage{booktabs}
\usepackage{xspace}
\usepackage{graphics}

\usepackage{cite}

\numberwithin{theorem}{section}

\newcommand{\M}[1]{\mathtt{#1}}
\newcommand{\V}[1]{\mathbf{#1}}

\newcommand{\diag}{\textrm{diag}}

\newcommand{\comment}[1]{}
\newcommand{\p}{\texttt{P}}
\newcommand{\g}{{\texttt{G}}}
\newcommand{\cam}[1]{{\mathcal{#1}}}

\def\Hk{Heikkil{\"{a}}\xspace}  
\def\gb{Gr{\"o}bner basis\xspace}

\def\pinhole{{P}}
\def\gen{{G}}

\newcommand{\best}[1]{\textcolor{red}{\textbf{#1}}}
\newcommand{\secondBest}[1]{\textcolor{blue}{\textbf{#1}}}

\definecolor{mygray}{gray}{.93}
\definecolor{mygray2}{gray}{.97}

\usepackage[pagebackref=true,breaklinks=true,letterpaper=true,colorlinks,bookmarks=false]{hyperref}

\newcommand{\PAR}[1]{\vskip4pt \noindent{\bf #1~}}

\iccvfinalcopy % *** Uncomment this line for the final submission

\def\iccvPaperID{} % *** Enter the ICCV Paper ID here

\ificcvfinal\fi

\begin{document}

%%%%%%%%% TITLE
\title{Calibrated and Partially Calibrated Semi-Generalized Homographies}
\author{Snehal Bhayani$^\textrm{1}$ \quad
Torsten Sattler$^\textrm{2}$ \quad
Daniel Barath$^\textrm{3}$ \quad
Patrik Beliansky$^\textrm{4}$ \\
Janne \Hk$^\textrm{1}$ \quad
Zuzana Kukelova$^\textrm{5}$\\
$^\textrm{1}$Center for Machine Vision and Signal Analysis, University of Oulu, Finland\\
$^\textrm{2}$Czech Institute of Informatics, Robotics and Cybernetics, Czech Technical University in Prague\\
$^\textrm{3}$Computer Vision and Geometry Group, Department of Computer Science, ETH Zürich\\
$^\textrm{4}$Faculty of Mathematics and Physics, Charles University, Prague\\
$^\textrm{5}$Visual Recognition Group, Faculty of Electrical Engineering, Czech Technical University in Prague
}

% \and

\maketitle
\ificcvfinal\thispagestyle{empty}\fi

%%%%%%%%% ABSTRACT
\begin{abstract}
\noindent  In this paper, we propose the first minimal solutions for estimating the semi-generalized homography given a perspective and a generalized camera. 
The proposed solvers use five 2D-2D image point correspondences induced by a scene plane.
One group of solvers assumes the perspective camera to be fully calibrated, while the other estimates the unknown focal length together with the absolute pose 
parameters. 
This setup is particularly important in structure-from-motion and visual localization pipelines, where a new camera is localized in each step with respect to a set of known cameras and 2D-3D correspondences might not be available.
%As a consequence of 
Thanks to
a clever parametrization and the elimination ideal method, 
%our approach leads to solvers,  which 
our solvers only need to solve a univariate polynomial of degree five or three, respectively a  system of polynomial equations in two variables. 
%in two unknowns.
% our approach only needs to solve a univariate polynomial of degree five or three. 
All proposed solvers are stable and efficient as demonstrated by a number of synthetic and real-world experiments.

\end{abstract}

%%%%%%%%% BODY TEXT
% \vspace{-12pt}
\section{Introduction}
\noindent 
Estimating the homography between two cameras observing a planar scene is a crucial problem in computer vision with applications, \eg, in structure-from-motion (SfM)~\cite{wu2013towards, snavely2006photo, sweeney2015optimizing, schonberger2016structure}, localization~\cite{Sattler2017PAMI,Brachmann2019ICCVa,Sarlin2019CVPR}, visual odometry~\cite{mur2015orb,mur2017orb}, camera calibration~\cite{Schops_2020_CVPR,zhang2000TPAMI}, and image retrieval~\cite{Philbin07CVPR,Weyand11ICCV}. It is one of the oldest camera geometry problems with many solutions including the well-known normalized direct linear transform (DLT) method~\cite{hartley2003multiple} for estimating the homography from a minimum of four point correspondences; the minimal solutions based on affine~\cite{koser2009geometric,barath2017theory} or SIFT correspondences~\cite{barath2018five,barath2019homography}; solutions assuming known gravity direction~\cite{saurer2017homography, ding2019efficient} 
%special type of motions~\cite{}\todo{}; 
or cameras with radial distortion~\cite{fitzgibbon2001simultaneous,hartley2003multiple,brown2007minimal,jin2008three,byrod2009minimal,kukelova2015radial}.
All %of the 
above-mentioned algorithms assume that both cameras satisfy the central perspective projection model (potentially, with radial distortion), \ie, they assume that both cameras have a single center of projection. 
Surprisingly, the problem of estimating a homography has not been studied for generalized cameras.

\begin{figure}[t]
    \centering
	    \includegraphics[width=0.58\columnwidth]{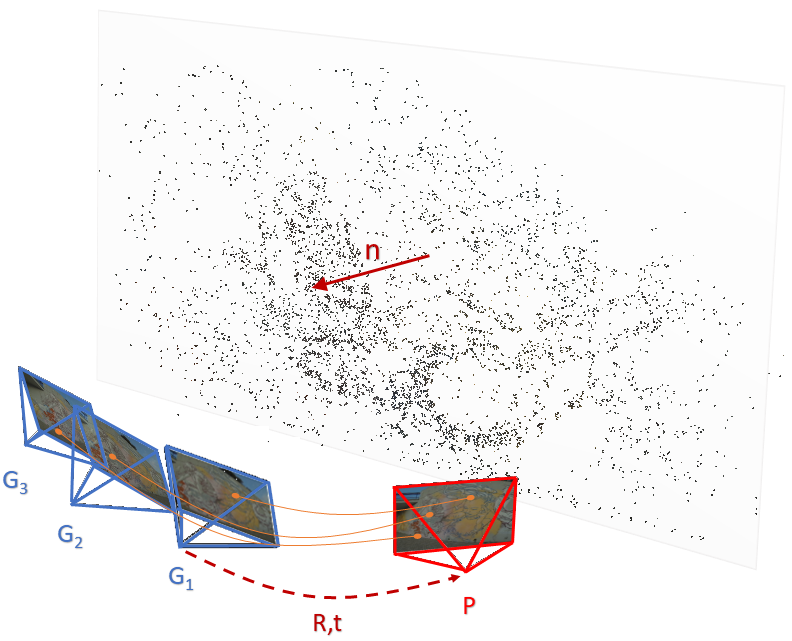}
        \caption{An illustration of the problem configuration. }
    \label{fig:illustration}
\end{figure}

A generalized camera~\cite{pless2004camera} is a camera that captures some arbitrary set of rays and does not adhere to the central perspective projection model. Such a camera model is practical and appears, \eg, in applications that exploit multi-camera configurations, like stereo-pairs, SfM~\cite{Wu_iccv15}, or in localization pipelines~\cite{Wald2020ECCV,Stenborg20203DV}. Such pipelines often are based on sequences of images, where there might be a set of cameras with known poses and we are given a new image which is to be registered to a generalized camera composed of the known perspective ones.
Estimating the camera pose, \wrt the generalized camera, in such situations often leads to results superior, in terms of accuracy, to considering only pair-wise epipolar geometries, especially thanks to a larger field-of-view of the generalized camera~\cite{sweeney2015computing}.
Also, it has the advantage of recovering the absolute pose~\cite{Wu_iccv15}, \ie, the scale of the translation, which is a severe deficiency of epipolar geometry-based relative pose estimation. 

While the problem of estimating the absolute pose of a generalized camera can be solved very efficiently~\cite{kukelova2016efficient}, \ie, there exists a solution that solves only 3 quadratic equations in 3 unknowns and runs in a few $\mu$s, the problem of estimating the relative pose of two generalized cameras is significantly more complex~\cite{Stewnius2005SolutionsTM}. This problem results in a system of 15 polynomial equations, each of degree $6$, with $64$ solutions. The final solver based on the \gb method~\cite{Stewnius2005SolutionsTM} is infeasible for real-time applications.

In~\cite{Wu_iccv15}, the authors considered a semi-generalized epipolar geometry problem, \ie, the problem of estimating the relative pose together with the scale of the translation between one perspective and one generalized camera from 2D-2D correspondences. In this paper, four minimal solvers were presented, \ie, $\mathbf{\M E_{5+1}}$ and $\mathbf{\M E_{4+2}}$ for calibrated pinhole cameras, and $\mathbf{\M Ef_{6+1}}$ and $\mathbf{\M Ef_{5+2}}$ for pinhole cameras with unknown focal length. 
Here, 4+2 denotes a configuration where four point correspondences come from one camera  
% The indices in the solver names denote the number of point correspondences coming from one camera 
$G_i$ of the generalized camera $G$ and the remaining two from 
%one or more 
one or two other cameras. The authors %demonstrated
showed the applicability of the proposed $\mathbf{\M E_{4+2}}$ and $\mathbf{\M Ef_{5+2}}$ solvers for incremental SfM %without 2D-3D point correspondences. %for incremental 
% camera registration 
in the absence of 2D-3D point correspondences. %, especially, 
% in an SfM pipeline. %structure-from-motion pipeline. 
However, the $\mathbf{\M E_{4+2}}$ and $\mathbf{\M Ef_{5+2}}$ solvers perform operations on large matrices and, thus, are impractical for real-time applications, with running times of $1.2 ms$ and $13.6 ms$, respectively.
%have rather large solvers and hence much slower s compare to the other two, 
The $\mathbf{\M E_{5+1}}$ and $\mathbf{\M Ef_{6+1}}$ solvers %, also mentioned in this paper, 
are based on the existing efficient five-point $\mathbf{\M E5}$~\cite{nister2004efficient} and the six point $\mathbf{\M E6f}$~\cite{bujnak20093d} relative pose methods. These solvers actually do not 
% gain 
%bring 
% any 
benefit from the generalized camera setup, except %the fact 
that one additional point correspondence is used 
to estimate %for the estimation of 
the scale of the translation. 
Moreover, the $\mathbf{\M E_{5+1}}$ and $\mathbf{\M Ef_{6+1}}$ solvers require $5$ ($6$ for unknown focal length) point correspondences to be detected by the same camera. 
This criterion may be problematic in the absence of enough inlier point matches. 
Note that~\cite{Wu_iccv15} does not solve all possible configurations of point correspondences that can appear  in the semi-generalized setup due to the complicated systems of polynomial equations. 
% Their solvers %proposed in~\cite{Wu_iccv15} 
% do not work if the 
Further, \cite{Wu_iccv15}  cannot handle generalized cameras %containing 
with more than three cameras and having fewer than four correspondences with all cameras. %come from a single camera. 

In this paper, we study a similar %camera 
setup as~\cite{Wu_iccv15}, \ie, one perspective and one generalized camera. 
However, we assume that these cameras observe a planar scene, see Fig.~\ref{fig:illustration}. 
We present the first minimal solutions for estimating the pose between a perspective and a generalized camera from 2D-2D correspondences induced by a plane, \ie, the first minimal solutions for the so-called {\em semi-generalized homography}. %problem. 
The proposed solvers use five 2D-2D image point correspondences %induced by a planar scene 
and assume either a calibrated or a perspective camera with unknown focal length. 
This setup is particularly important in SfM and localization, where a new camera is localized with respect to a set of known ones and 2D-3D correspondences might not be available, \eg, due to memory restrictions or to avoid matching features between individual cameras in the generalized camera. % or we do % \begin{itemize}[leftmargin=*]

% not want to do expensive feature matching between individual cameras in the generalized camera. 

\noindent The main {\bf contributions} of the paper are as follows:
% \vspace{-0.8ex}
% \begin{itemize}[leftmargin=*]
% \item 
\noindent \textbf{1)} A theoretical analysis of the {\bf new semi-generalized homography} problem for calibrated and partially calibrated cameras 
%with unknown focal length 
and a formulation of the problem as a system of linear equations in twelve unknowns.

% \vspace{-0.8ex}
% \item 
\noindent \textbf{2)} Derivation of {\bf new  constraints} for the semi-generalized homography using the elimination ideal theory~\cite{kukelova2017clever}. 

% \vspace{-0.8ex}
% \item
\noindent \textbf{3)} {\bf A class of efficient minimal solvers for calibrated cameras}, $\mathbf{s\M{H}5_2}$, $\mathbf{s\M{H}5_3}$, $\mathbf{s\M{H}5_4}$, $\mathbf{s\M{H}4.5_2}$ and $\mathbf{s\M{H}4.5_3}$  
that only need to solve a $5^{th}$ ($3^{rd}$) degree univariate polynomial, a linear system, or a system of equations in two unknowns. %\vspace{-0.8ex}

% \item
\noindent \textbf{4)} {\bf Two new efficient minimal solvers for partially calibrated cameras} $\mathbf{s\M{H}5f_2}$ and $\mathbf{s\M{H}5f_3}$ that need to solve a univariate polynomial of degree five (three). 

% \vspace{-0.8ex}
% \item
\noindent \textbf{5)} Our %The proposed 
solvers do not need 3D points or 2D-2D matches between individual cameras from the generalized camera and cover all scenarios where it is possible to estimate the scale. 

% \vspace{-0.8ex}
% \item 
\noindent \textbf{6)} Compared to~\cite{Wu_iccv15}, our solvers cover {\bf all possible minimal configurations of point correspondences} as well as numbers of cameras in the generalized camera. 
Code is available at \href{https://github.com/snehalbhayani/SemiGeneralizedHomography}{github.com/snehalbhayani/SemiGeneralizedHomography}. 

% \vspace{-0.8ex}
% \item 
% \noindent \textbf{6)}  A software implementation of our proposed solvers, and the scripts for evaluating the same are available at~\cite{SemiGenHom}.
% \end{itemize}
% Our implementation is available at \href{https://github.com/snehalbhayani/SemiGeneralizedHomography}{github.com/snehalbhayani/SemiGeneralizedHomography}.

\section{Problem Formulation}
\noindent 
First, we set up notations and conventions that we will follow for the rest of the paper. Let $\cam{\pinhole}$ denote the perspective camera, while the generalized camera is denoted as $\cam{G}$. We assume that the generalized camera $\cam{G}$ is fully calibrated, and it consists of a set of perspective cameras $\lbrace \cam{\gen}_1, \cam{\gen}_2, \dots \cam{\gen}_k \rbrace$.
For the pinhole camera $\cam{\pinhole}$, we consider two different cases, \ie, the case where $\cam{\pinhole}$ is fully calibrated, and the case when its calibration matrix is of the form $\M{K} = \diag(f,f,1)$ with unknown focal length $f$. 

In the following text, we will consider several different coordinate systems, \ie, the global coordinate system, the local coordinate system of the perspective camera $\cam{\pinhole}$, and local coordinate systems of perspective cameras $\cam{\gen}_i$. Let $\M{R}_{\g_i}$, $\V{t}_{\g_i}$ and $\M{R}_{\p}$, $\V{t}_{\p}$ denote the rotations and translations required to align local the coordinate systems of $\cam{\gen}_i$ respectively $\cam{\pinhole}$, to the global coordinate system.
Without loss of generality, we can assume that the global coordinate system coincides with the local coordinate system of $\cam{\gen}_1$, \ie, $\M{R}_{\g_1} = \M{I}$ and $\V{t}_{\g_1}= [0,0,0]^\top$. Sometimes we will call this system the local coordinate system of the generalized camera $\cam{G}$. Therefore, the global coordinate system, the local coordinate system of $\cam{\gen}_1$, and the local coordinate system of $\cam{G}$ are interchangeable.
We will use the upper index to denote the coordinate system. 
For example, $\V X^{\p} \in \mathbb{R}^3$ and  $\V X^{\g} \in \mathbb{R}^3$ are the coordinates of the point $\V X$ in the local coordinate system of $\cam{\pinhole}$ and the local coordinate system of $\cam{G}$, respectively, and it holds that $\V X^{\g} = \M R_{\p}\V X^{\p} + \V{t}_{\p}$.

Our objective is to estimate the rotation $\M R \in \bf{SO}(3)$ and translation $\V t \in \mathbb{R}^3$ between the perspective camera $\cam{P}$ and the generalized camera $\cam{G}$, \ie, the rotation and translation that align the local coordinate system of $\cam{P}$ to that of $\cam{G}$. 
%Note that $\M R = \M R_\p$  and $\V t = \V t_\p$.
Note that, since the local coordinate system of 
%the generalized camera
$\cam{G}$ coincides with the global coordinate system,
$\M R = \M R_\p$  and $\V t = \V t_\p$.

For the estimation of $\M{R}$ and $\V{t}$, we will use 2D-2D correspondences detected between $\cam{P}$ and the cameras in $\cam{\gen}$. We assume that these point correspondences are projections of co-planar 3D points $\V X_j$ satisfying: $\V{n}^\top \V X_j + d = 0$, where $\V{n} \in \mathbb{R}^3$ denotes the normal of the scene plane $\pi$ and $d$ denotes the plane intercept.
Note that the same plane can be defined by the normal $\widetilde{\V{n}} = \V{n}/d$ and the equation $\widetilde{\V{n}}^\top\V X_j + 1 = 0$.

%We can consider the quantity $\V{n}/d$ to define the same plane in the text and by abuse of notation, we will henceforth use $\V{n}$ instead of $\V{n}/d$.

\subsection{Semi-Generalized Homography}
\noindent 
% Let us assume 
Consider a 3D point $\V{X}_j$ observed by the perspective camera $\cam{P}$ and the camera $\cam{G}_i$, \ie, the  $i$-th constituent perspective camera from the generalized camera $\cam{G}$.
Let us denote the image points detected in $\cam{P}$ and $\cam{G}_i$ as $\V p_j = [x_j, y_j, 1]^\top$ and $\V g_{ij} = [x^{G_{i}}_j, y^{G_{i}}_j, 1]^\top$, respectively. 
With this notation, the coordinates of the 3D point $\V{X}_j$ in the local coordinate system of $\cam{P}$ can be expressed as 
\begin{eqnarray}
\V{X}_j^{\p} = \alpha_{j} \M{K}^{-1} \V p_j \enspace, \label{eq:Xp}
\end{eqnarray}
where $\M{K}$ is the calibration matrix of the camera $\cam{P}$ and $\alpha_j$ represents the depth of the point $\V X_j$ in $\cam{P}$. A similar relationship holds for the coordinates of the 3D point $\V X_j$ in the local coordinate system of $\cam{\gen}_i$,
\begin{eqnarray}
\V{X}_j^{\g_i} = \beta_{ij} \M{K}_{\g_i}^{-1} \V g_{ij} \enspace, \label{eq:XGi}
\end{eqnarray}
where $\M{K}_{\g_i}$ is the calibration matrix of the camera $\cam{\gen}_i$ and $\beta_{ij}$ represents the depth of the point $\V X_j$ in $\cam{\gen}_i$.

To obtain the relationship between $\V{X}_j^{\p}$ and $\V{X}_j^{\g_i}$, we have to transform them into the same coordinate system, \ie, in this case the global coordinate system. This gives us the following constraint
\begin{eqnarray}
\alpha_{j}  \M R \M{K}^{-1} \V p_j +\V t = \beta_{ij} \M{R}_{\g_i}  \M{K}_{\g_i}^{-1} \V g_{ij} + \V t_{\g_i} \enspace.
\end{eqnarray}
Note that here we use the fact that $\M R = \M R_\p$  and $\V t = \V t_\p$.
Since in our case $\M{R}_{\g_i}, \V{t}_{\g_i}$ and  $\M{K}_{\g_i}^{-1}$ are known, we will, for better readability, substitute  $\V q_{ij} = \M{R}_{\g_i}  \M{K}_{\g_i}^{-1} \V g_{ij}$ and obtain
\begin{eqnarray}
\alpha_{j}  \M R \M{K}^{-1} \V p_j +\V t = \beta_{ij} \V q_{ij} + \V t_{\g_i} \enspace. \label{eq:global_q}
\end{eqnarray}
The 3D point $\V X_j$ is lying on the plane $\pi$, \ie, $\V{X}_j^{\p}$  should satisfy $(\widetilde{\V{n}}^{\p})^\top\V X_j^\p + 1 = 0$, where $\widetilde{\V{n}}^{\p} \in \mathbb{R}^3$ is the normal of the plane $\pi$ in the local coordinate system of $\cam{P}$. For simplicity, in the rest of the text, we will omit the upper index $\p$ in $\widetilde{\V{n}}^{\p}$.
The depth $\alpha_{j}$ in~\eqref{eq:Xp} can be then expressed using the normal $\widetilde{\V{n}}$ as 
\begin{equation}
    \alpha_{j} = \dfrac{-1}{\widetilde{\V{n}}^\top \M{K}^{-1} \Vec{\V{p}_j}} \enspace.
    \label{eq:alpha}
\end{equation}
Let us consider a $3 \times 3$ homography matrix $\M{H}$ of the form $\M{H} = \M{R} - \V{t} \widetilde{\V{n}}^\top$. By substituting~\eqref{eq:alpha} into~\eqref{eq:global_q} we obtain   
\begin{eqnarray}
\alpha_{j}  \M H \M{K}^{-1} \V p_j = \beta_{ij} \V q_{ij} + \V{t}_{\g_i} \label{eq:global_H} \enspace .
\end{eqnarray}
%
%We call this equation the semi-generalized homography constraint.
Eq.~\eqref{eq:global_H} is the  basic 
semi-generalized homography constraint.
The depths $\beta_{ij}$ can be easily eliminated from this constraint~\eqref{eq:global_H} by
multiplying it with the skew-symmetric matrix $[\V{q_j}]_\times$ from the left side, resulting in \begin{eqnarray}
[\V{q}_{ij}]_\times(\alpha_{j}  \M H \M{K}^{-1} \V p_j - \V t_{\g_i})=\V 0 \label{eq:skewH} \enspace.
\end{eqnarray}
Let us denote $\M{G} = \M{H} \M{K}^{-1}$. By dividing~\eqref{eq:skewH} with $\alpha_j$ and using~\eqref{eq:alpha}, we obtain the equations
\begin{equation}
[\V{q}_{ij}]_\times(\M G \V p_j + (\V{m}^\top {\V{p_j}})\V t_{\g_i})=\V 0 \label{eq:skewGM} \enspace,
\end{equation}
where $\V{m} = \M{K}^{-1} \widetilde{\V{n}}$. 
Note that we are able to eliminate  
the unknown depths $\alpha_j$ from~\eqref{eq:skewH}, and to derive simple linear constraints~\eqref{eq:skewGM}  for the semi-generalized homography thanks to the special parameterization~\eqref{eq:alpha}, based on the normal 
%of the plane 
$\widetilde{\V{n}}^{\p}$
expressed in the coordinate system of $\cam{P}$.
%Note that this special parameterization, based on the normal of the plane expressed in the coordinate system of $P$, allowed us to eliminate  
%the unknown depths $\alpha_j$ from~\eqref{eq:skewH}, and to derive simple linear constraints in 12 unknowns~\eqref{eq:skewGM} for the semi-generalized homography. 
Each 2D-2D correspondence $\V{p}_j \leftrightarrow \V{q}_{ij}$  gives us three equations 
%in 12 unknowns 
of the form~\eqref{eq:skewGM}, from which only two are linearly independent.

%%-----------------------
\subsection{Semi-Generalized Homography Constraints}
\noindent
Besides the constraints~\eqref{eq:skewGM} induced by a 2D-2D correspondence $\V{p}_j \leftrightarrow \V{q}_{ij}$, there are other ones arising from the form of the matrix $\M G$.
The constraints~\eqref{eq:skewGM} are linear in the 12 unknowns, \ie, elements of the matrix $\M G$ and the vector $\V m$. However, $\M G$ and $\V m$ are not independent since $\M G = \M{H}\M{K}^{-1} = \M{R}\M{K}^{-1}  - \V{t}\V{m}^\top $. Moreover, the rotation matrix $\M R \in \bf{SO}(3)$ introduces additional constraints. All these constraints, \ie, constraints originating from the form
\begin{eqnarray}\label{eq:constraints_for_I}
\M G - \M{R}\M{K}^{-1}  + \V{t}\V{m}^\top = \M{0}, \quad
\M{R}^\top \M{R} = \M{R} \M{R}^\top = \M{I}_{3\times 3} \enspace , 
\end{eqnarray}
can be used to define an ideal $I \subset \mathbb{C}\left[ \varepsilon \right] $~\cite{cox2006using}, where $\varepsilon$ contains 9 unknowns from $\M G$, 9 from $\M R$, 3 from $\V t$, 3 from $\V m$ and the inverse of the focal length $w= {1 \over f}$. Now we can use the elimination ideal technique~\cite{kukelova2017clever} to eliminate 9 unknowns of $\M R$, 3 of $\V t$, and $w$ from this ideal. 
\Ie, we compute an elimination ideal $I_1$ that will contain only polynomials in 12 unknowns from $\M G$ and $\V m$. Eq.~\eqref{eq:constraints_for_I} does not include the constraint $det(\M{R}) = 1$, as it does not change $I_1$, whose generators are the same for $\M{G}$ and $-\M{G}$, \ie, both $\M{R}, \V{t}$ and $-\M{R}, -\V{t}$ are valid solutions. This ambiguity is resolved at a later stage when decomposing the homography matrix $\M{H}$. %Note that t
The elimination ideal $I_1$ can be computed offline using some algebraic geometry software like Macaulay 2~\cite{M2}. We found that such an elimination ideal is generated by 4 polynomials (three of degree 3 and one of degree 4) in 12 unknowns. Similarly, for the calibrated case, \ie, \ $\M K = \M I$ there are 10 such generators of $I_1$. For more details on elimination ideals we refer to~\cite{cox2006using,kukelova2017clever}. 

These 10 new constraints (4 for the unknown focal length case) in 12 unknowns from $\M G$ and $\V m$ together with the linear equations~\eqref{eq:skewGM} in these unknowns can be used to solve for $\M R, \M t$ (and $f$). After a null-space re-parameterization of $\M G$ and $\V m$ using the linear equations~\eqref{eq:skewGM} for 5 point matches, we can transform these equations to 10 (4) polynomial equations in 2 unknowns, respectively 3 unknowns for the 4.5 point matches required for the calibrated case. Such systems can be solved, \eg, using the automatic generator of \gb solvers~\cite{kukelova2008automatic, larsson2017efficient} 
%This results in  performing Gauss-Jordan (G-J) elimination of a $12 \times 28$, respectively $33 \times 49$ matrix and Eigenvalue decomposition of a $16 \times 16$ matrix for the calibrated case and G-J elimination of a $6 \times 12$ matrix and Eigenvalue decomposition of a $6 \times 6$ matrix for the unknown focal length case. The solvers return up to $16$ (calibrated), respectively $6$ (unknown $f$), real solutions to $\M G$ and $\V m$. 
and they return up to $16$ (calibrated), respectively $6$ (unknown $f$), real solutions to $\M G$ and $\V m$. For details on solvers sizes see the supp. material (SM).

In order to make the solvers more efficient, we introduce an additional change of variables by assuming that one of the elements of $\M G$ is non-zero, \eg, $g_{33} \neq 0$. This assumption can introduce a degeneracy. However, such a degeneracy is not crucial in practical applications and can be avoided as %it was 
discussed in~\cite{chum2010accv}. Moreover, our change of coordinate system used for the calibrated solver directly avoids this degeneracy. The situation for the focal length case is discussed in more detail in %the Supp. material (SM).
the SM.

With this assumption, we introduce new variables $g_{kl}^\prime = {g_{kl} \over g_{33}}$, for $ \forall kl  \neq 33$ and $m^{\prime}_k = {m_{k} \over g_{33}}$, $k = 1, 2,3$, where $g_{kl}$ are elements from the $k^{th}$ row and $l^{th}$ column of the matrix $\M G$ and $m_{k}$ are elements of the vector $\V m$. 
The variable change is also applied to the 10 (4) generators of $I_1$, defining a new ideal $I_1^{\prime}$. Using again the elimination ideal technique~\cite{kukelova2017clever}, we can eliminate $g_{33}$ from the ideal $I_1^{\prime}$, leading to a new ideal $I^{\prime}_{2}$.
For the calibrated case, the ideal $I^{\prime}_{2}$ is generated by five polynomials $e_i$, each of degree $5$, %and 
in the $8+3$ unknowns, $g_{kl}^\prime$ for $ \forall kl  \neq 33$ and $m^{\prime}_k$, $k = 1,2,3$. For the unknown focal length case, it is generated by only a single polynomial $e$ of degree five.
%in these $8+3$ unknowns.
More details on the form of the generators of the elimination ideals $I_1$ and $I_2^\prime$ for both the cases, together with the input code for Macaulay2 used to compute these generators, is provided in the SM.
Note that the derivation of these $5^{th}$ degree constraints
%$f$
as well as the above mentioned generators of $I_1$ is crucial for the efficiency of the final solvers. Without using the elimination ideal tricks, one would need to work directly with the parameterization of $\M G$ using the rotation matrix, leading to complex systems of polynomial equations in many unknowns and with huge solvers. Next we show how to solve these equations for a calibrated $\cam{P}$ and then for the case when the focal length of $\cam{P}$ is unknown. 

%%-----------------------

\begin{table*}[t!]
\setlength{\tabcolsep}{5pt}
\scriptsize
	\newcommand{\tabincell}[2]{\begin{tabular}{@{}#1@{}}#2\end{tabular}}
	\begin{center}
		\begin{tabular}{lcccc>{\columncolor{mygray2}}c>{\columncolor{mygray2}}c>{\columncolor{mygray}}c>{\columncolor{mygray}}c >{\columncolor{mygray}}c >{\columncolor{mygray}}c>{\columncolor{mygray}}c>{\columncolor{mygray}}c>{\columncolor{mygray}}c}
			\toprule
			 & $\mathbf{\M E_{5+1}}$ & $\mathbf{\M E_{4+2}}$ & $\mathbf{\M Ef_{6+1}}$ & $\mathbf{\M Ef_{5+2}}$&  P3P+N  & P5Pf+N & $\mathbf{s\M{H}5_2}$& $\mathbf{s\M{H}4.5_2}$ & $\mathbf{s\M{H}5_3}$ & $\mathbf{s\M{H}4.5_3}$ & $\mathbf{s\M{H}5_4}$ & $\mathbf{s\M{H}5f_2}$  & $\mathbf{s\M{H}5f_3}$\\
			\midrule
			Ref & \cite{Wu_iccv15} & \cite{Wu_iccv15} &\cite{Wu_iccv15} & \cite{Wu_iccv15} &   &  &  &  &  &    &  & & \\
			Focal &  & &\checkmark & \checkmark&  & \checkmark &    &  & & & & \checkmark& \checkmark\\
		   $\#$ pt & 6 & 6& 7& 7& 6 & 8 &  5& 4.5  & 5& 4.5 & 5 & 5& 5\\
		   $\#$ sol & 10 & 40 & 10 & 50 & 4 & 4 & 5 & 16   & 3& 12& 1 & 5 & 3 \\
		    \midrule
		     &  & & & & & Complexity& & & & &  & & \\
		    \midrule
			G-J/LU & $10\times20$ & $73\times113$ & $11\times20$ &$378\times428$& &$5\times8$&$8\times10$&$11\times27$ &$8\times10$&$23\times35$ & $8\times9$ & $8\times10$&$8\times10$\\
% 			SVD &&&$3\times3$&&&$3\times3$&&&&\\
			Eigen &&$40\times40$&$9\times9$&$50\times50$&$3\times3$&&&$16 \times 16$&&$12 \times 12$&&&\\
 			Sturm &10&&&&&&5&&3&&&5&3\\
 			QR & $5\times9$&&&&&$5\times8$&&&&&&&\\
			%in p5pf  A.colPivHouseholderQr().solve(B); a...5x2,b...5x1
			%6+1 computing nullspace of 6x9 matrix
			\bottomrule
		\end{tabular}
	\end{center}
	\caption{Comparison of the proposed solvers (gray) vs. the state-of-the-art.}
	\label{table:computationalcomplexity}
	\normalsize
\end{table*} 

\subsection{Calibrated Camera Solvers}
\noindent In this case, we can assume that $\M K = \M{I}_{3 \times 3}$, leading to $\M{G} = \M{H} = \M{R} - \V{t} \widetilde{\V{n}}^\top$. Here, we have $9$ DOF, $3$ for each $\M{R}$, $\V{t}$ and $\widetilde{\V{n}}$. Each point correspondence leads to $2$ linearly independent homogeneous constraints of the form~\eqref{eq:skewGM} and thus $4.5$ correspondences are sufficient to solve this problem. However we still need to sample $5$ point correspondences, $\V{p}_j \leftrightarrow \V{q}_{ij}$, $j=1, \dots 5$, resulting in 
%a total of 
$10$ constraints. 

There are two ways to deal with this over-constrained formulation:  
%of which 
1) 
% One is to 
use all $10$ constraints from the $5$ point correspondences and only %consider 
one constraint from the $5$ generators $e_i$ of the ideal $I^\prime_2$.
%on the $\M G$ matrix. 
2) 
% Another way is to 
use $9$ constraints by considering only $4.5$ point correspondences, and instead use all $5$ constraints $e_i$ on the $\M G$ matrix\footnote{Note that the x- and y-coordinates of all 5 correspondences are used. However, one constraint from~\eqref{eq:skewGM} originating from the matches is omitted.}. 
The first approach results in solvers $\mathbf{s\M{H}5_2}$ and $\mathbf{s\M{H}5_3}$ that have to find the roots of a univariate polynomial. The second approach leads to solvers $\mathbf{s\M{H}4.5_2}$ and $\mathbf{s\M{H}4.5_3}$ that have to solve a system of 5 equations in 2 unknowns. The $4.5$ point solvers are slightly slower than the $5$ point solvers. %, h
However, since they use all constraints on $\M G$, they result in a correctly decomposable homography and therefore smaller errors in the presence of noise. These solvers are described in detail in the SM.
Next we describe the $5$ point solvers, $\mathbf{s\M{H}5_2}$ and $\mathbf{s\M{H}5_3}$.
%In the first case, we have a univariate solver, with $5$ solutions. This solver may have instability in the presence of noise. However in the second case, we have a slightly slower solver with more solutions, but much better stability in the presence of noise. We describe calibrated solvers $\mathbf{s\M{H}5_2}$ and $\mathbf{s\M{H}5_3}$ based on the first approach in this paper, and provide details of the solvers, $\mathbf{s\M{H}4.5_2}$ and $\mathbf{s\M{H}4.5_3}$, based on the second approach in the Supplementary material.

% leading to a single-variable parameterization of the unknowns in $\varepsilon^{\prime}$ but using only one polynomial out of the five polynomial generators of the elimination ideal $I_2^{\prime}$. This leads to a fast but unstable solver in the presence of noise. Alternately, we can use only $4.5$ point correspondences, leading to $\M C$ to be of size $7 \times 10$ and hence to a 2-variable parameterization of the unknowns in $\varepsilon^{\prime}$. 

Without loss of generality (w.l.o.g.), we assume that the first point correspondence is observed in camera $\cam{G}_1$, \ie, $i = 1$ where $\V {t}_{\g_1} = [0,0,0]^\top$, and pre-rotate the local coordinate systems of $\cam{P}$ and $\cam{G}_1$ such that $\V{p}_1 = [0,0,1]^\top$ and $\V{q}_{11} = [0,0,1]^\top$. This simplifies the equations and after substituting into~\eqref{eq:skewGM}, we have $g_{13}=0,\;g_{23}=0$.  Moreover, we can safely assume $g_{33} \neq 0$ and divide these equations by $g_{33}$, transforming them into non-homogeneous equations in 
%same 
 $9$ unknowns, $\varepsilon^{\prime} = \lbrace g^{\prime}_{11}, g^{\prime}_{12}, g^{\prime}_{21}, g^{\prime}_{22},   g^{\prime}_{31}, g^{\prime}_{32}, m^{\prime}_1, m^{\prime}_2, m^{\prime}_3 \rbrace$. The remaining $4$ point correspondences lead to $8$ linearly independent equations that can be written in %a 
 matrix form %as:
\begin{equation}
    \M{C} \V{b} = 0 \enspace,
    \label{eq:linearized2d2dconstraint}
\end{equation}
where $\M{C}$ is a ${8 \times 10}$ coefficient matrix and $\V{b}$ is a ${10 \times 1}$ vectorized form of the set of $\varepsilon^{\prime} \cup \lbrace 1 \rbrace$.  Next, we consider three cases based on the maximum number of correspondences coming from one camera $\cam{G}_i$.

% \vspace{-3.9mm}
\PAR{$\bf{s\M{H}5_2}$ solver:}
In this case no more than $2$ correspondences come from the same camera $\cam{G}_i$. Hence the matrix $\M{C}$ in~\eqref{eq:linearized2d2dconstraint} has a two dimensional null-space 
$\lbrace \V{b_1}, \V{b_2} \rbrace$. A solution to $\V{b}$ can be obtained as a linear combination $\V{b} = \gamma_1 \V{b_1} + \gamma_2 \V{b_2}$. Using %the constraint 
$b_{10} = 1$, we can express $\gamma_2$  as a linear polynomial in $\gamma_1$. Hence the variables in $\varepsilon^{\prime}$ can be parameterized as linear polynomials of %only one variable 
$\gamma_1$. This parameterization can be substituted into the generators of the ideal $I_2^{\prime}$ for the calibrated case. This leads to $5$ univariate polynomials $e_i(\gamma_1)$, each of degree $5$. % (w.l.o.g.), 
We choose one of these polynomials, which we solve using Sturm sequences~\cite{Hook1990UsingSS}. This results in up to five real solutions to~$\varepsilon^{\prime}$.

Next, we extract solutions to $g_{33}$. Writing $\M{G} = \M{H} = \M{R} - \V{t} \V{m}^\top$, we obtain a set of polynomial constraints. By variable elimination and substitutions, we obtain $1$ solution to $g_{33}$, unique up to a sign, which is fixed by constraining the solution of the plane vector $\V{n}$ so that the corresponding 3D point in $\cam{\pinhole}$ is in the front of the camera.  Solutions to $\M{G}$ as well as $\V{m}$ can be extracted from the solutions to $\varepsilon^{\prime}$ and $g_{33}$. We then decompose $\M H$ such that $det(\M{R}) = 1$ to obtain a set of relative poses $\M{R}$ and $\V{t}$.
%, from which we choose the feasible ones. 
Note that the constraints $e_i$ that were not used to obtain the solutions can be used to eliminate infeasible solutions.

% We note that choosing only one out of these five generators leads to solver instability in the presence of noise, as the generalized homography is not fully constrained. Hence, we propose a more stable variant by considering only $4.5$ point correspondences resulting in $7$ equations instead of $8$. This means that the matrix $\M C$ in~\eqref{eq:linearized2d2dconstraint} is of size ${7 \times 10}$, with a three dimensional null-space. Hence by following the same approach, a solution to $\V{b}$ can be obtained as a linear combination, $\V{b} = \gamma_1 \V{b_1} + \gamma_2 \V{b_2} + \gamma_3 \V{b_3}$. Assuming $b_{10} = 1$, $\gamma_3$ can be expressed as a linear polynomial in $\gamma_1$ and $\gamma_2$. Thus we have a two-variable parameterization of the unknowns in $\varepsilon^{\prime}$. The subsequent substitution into the generators of the ideal $I_2^{\prime}$ gives us $5$ polynomials in $2$ variables.

% \vspace{-3.9mm}
\PAR{$\bf{s\M{H}5_3}$ solver:}
If there are $3$ 2D-2D point correspondences %coming 
from the same camera $\cam{G}_i$, the situation is a bit different. 
 Let us assume, w.l.o.g., %without loss of generality, 
 that the points $\V{q}_{i2}$ and  $\V{q}_{i3}$  are observed in camera $\cam{G}_1$, \ie, $i = 1$. This is the same camera that observed the point $\V{q}_{11}$. The remaining two points can be observed by one camera $\cam{G}_j \neq \cam{G}_1$ or by two different cameras $\cam{G}_j ~\neq \cam{G}_k \neq \cam{G}_1$.
  In this case, G-J elimination of the matrix $\M{C}$ in~\eqref{eq:linearized2d2dconstraint}  leads to a matrix of a special form 
 \begin{eqnarray}
 \begin{bmatrix} \M{I}_{6 \times 6} & \M{0}_{6 \times 2} & \V{0}_{6 \times 1} & \V{c}_{6 \times 1} \\
 \M{0}_{2 \times 6} & \M{I}_{2 \times 2} & \V{d}_{2 \times 1} & \V{e}_{2 \times 1}
 \end{bmatrix} \V{b} = \V{0} \enspace ,
 \label{eq:H3GJ}
 \end{eqnarray}
 where the indices of the matrices and vectors indicate their sizes.  
 Since $\V b = [g^{\prime}_{11}, g^{\prime}_{12}, g^{\prime}_{21}, g^{\prime}_{22},   g^{\prime}_{31}, g^{\prime}_{32}, m^{\prime}_1, m^{\prime}_2, m^{\prime}_3, 1]^\top$, the first six rows of~\eqref{eq:H3GJ} directly give us  a solution to $g^{\prime}_{kl}$. The last two rows can be used to express $m^{\prime}_{1}, m^{\prime}_{2}$ as a linear function of $m^{\prime}_{3}$. Substituting this parameterization into the generators of $I^{\prime}_{2}$ yields five univariate polynomials, each of degree three. We can obtain up to three real solutions to $\varepsilon^{\prime}$ by solving one of these polynomials.
 %using Sturm sequences~\cite{Hook1990UsingSS}. 
 The remaining steps are similar to the $s\M H5_2$ solver.
 
% \vspace{-3.9mm}
\PAR{$\bf{s\M{H}5_4}$ solver:}
In this case, %we have 
$4$ points come %coming 
from the same camera $\cam{G}_i$. Let us assume, w.l.o.g., % without loss of generality, 
that the points $\V{q}_{i2}$, $\V{q}_{i3}$ and $\V{q}_{i4}$ are observed in camera $\cam{G}_1$, \ie, $i = 1$ and the fifth point $\V{q}_{25}$ is observed by camera $\cam{G}_2 \neq \cam{G}_1$. We estimate the semi-generalized homography $\M G$ by considering it as standard homography estimation problem from $4$ point correspondences~\cite{hartley2003multiple}. Decomposing $\M G = \M{R}  - \V{t}\V{m}^\top $ we obtain the rotation matrix $\M R$ and the translation vector $\V t$ up to scale. %, up to a non-zero scale. 
This scale can be computed from the constraint of the form~\eqref{eq:global_q}, induced by the fifth %point 
correspondence $\V{p}_5 \leftrightarrow \V{q}_{25}$, as
\begin{eqnarray}
\text{scale} = \dfrac{\V t_{\g_2}^\top [\M{R}\V{p}_5]_{\times} \V{q}_{25}}{\V{t}^\top [\M{R}\V{p}_5]_{\times} \V{q}_{25}} \enspace.
% \alpha_{5}  \M R \V p_5 +\V t = \beta_{25} \V q_{25} + \V t_{\g_2}.
\end{eqnarray}

\subsection{Unknown Focal Length Solvers}\label{sec:unknownfocalsolvers}
\noindent
In this case, we assume %the 
an unknown focal length $f$ in the calibration matrix $\M{K} = \diag(f,f,1)$. Therefore we have $10$ DOF
and we need five full 2D-2D correspondences to solve this problem.
%Hence unlike the case of calibrated cameras, this is a well-constrained formulation.
%
% Without loss of generality, we can assume that the first point correspondence is observed in camera $G_1$, \ie, $i = 1$. 
% For $i=1$, we have $\V {t}_{\g_1} = [0,0,0]^\top$. Moreover, without loss of generality we can pre-rotate the local coordinate systems of $P$ and $G_1$ such that $\V{p}_1 = [1,0,1]^\top$ and $\V{q}_{11} = [0,0,1]^\top$. 
% This simplifies the equations, and after substituting into~\eqref{eq:skewGM}, we have 
% \begin{eqnarray}
% g_{13}=-g_{11},\;g_{23}=-g_{21}.
% \label{eq:focal_G}
% \end{eqnarray}
% The remaining $4$ 2D-2D correspondences give us $8$ linearly independent constraints of the form~\eqref{eq:skewGM} which are homogeneous in the elements of $\M G$ and $\V m$. Similar to the calibrated case, these $8$ polynomials are de-homogenized by a variable substitution, leading to $8$ polynomials in $9$ unknowns. %$\varepsilon^{\prime} = \lbrace g^{\prime}_{11}, g^{\prime}_{12}, g^{\prime}_{21}, g^{\prime}_{22},   g^{\prime}_{31}, g^{\prime}_{32}, m^{\prime}_1, m^{\prime}_2, m^{\prime}_3 \rbrace$. 
%
Based on the maximum number of correspondences coming from one camera $\cam{G}_i$, we have two solvers: one where there are up to $2$ points from the same camera, $\mathbf{s\M{H}5f_2}$, and one where there are $3$ points from the same camera, $\mathbf{s\M{H}5f_3}$. 
These solvers solve univariate polynomials of degree $5$, respectively $3$, and they follow similar steps as those of the calibrated solvers $\bf{s\M{H}5}_2$ and $\mathbf{s\M{H}5}_3$. For more details on these solvers see the SM. 
 Note that for the case of $4$ points coming %are coming 
 from the same camera $\cam{G}_i$, one additional correspondence from camera $\cam{G}_j$ will not add sufficient constraints to recover both the unknown focal length and the unknown scale, which is proved in the SM.

\section{Experiments}
\begin{figure*}[t]
\centering
    \subfloat[]{\includegraphics[width=0.245\linewidth]{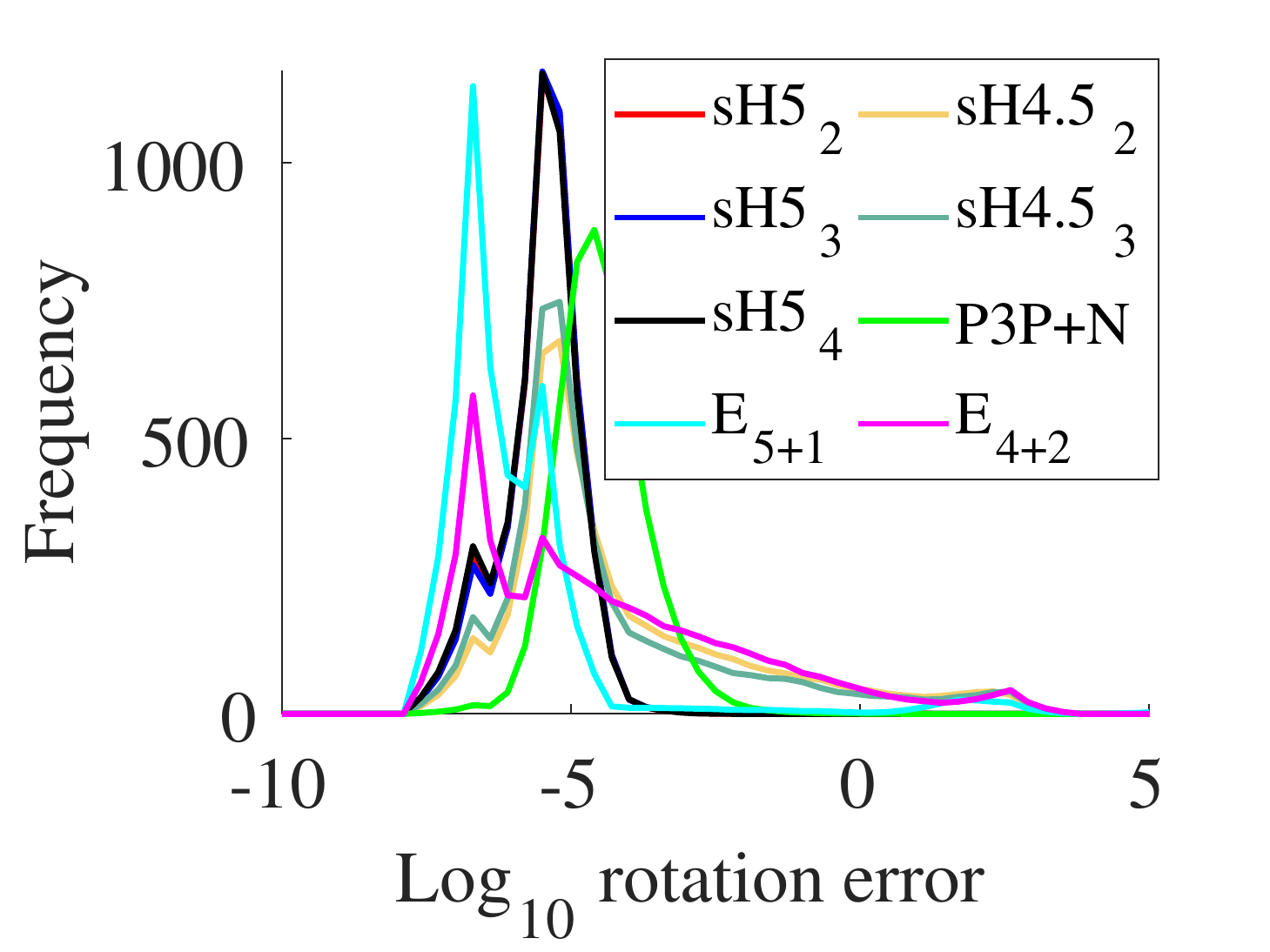}}
    \subfloat[]{\includegraphics[width=0.245\linewidth]{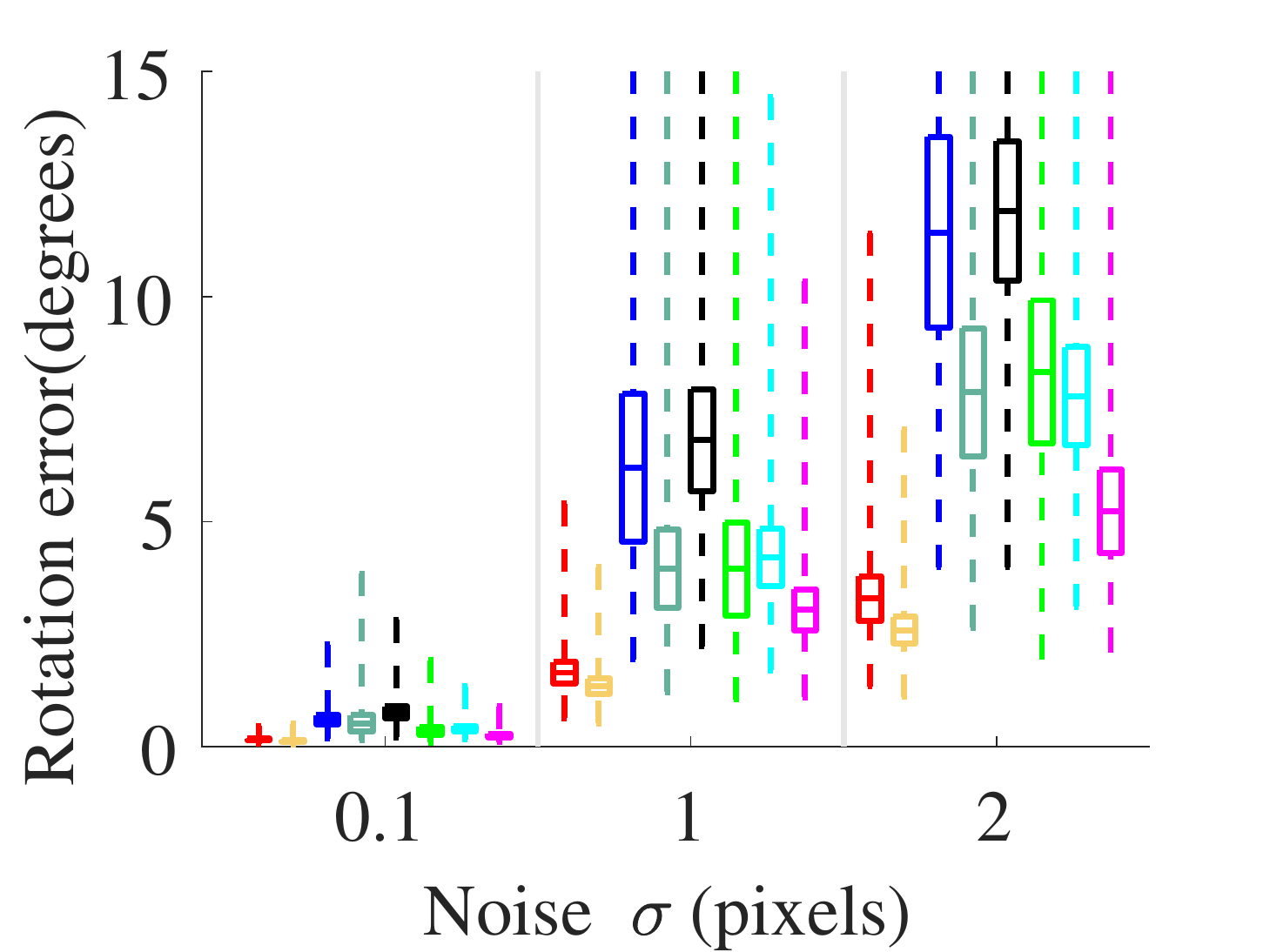} }
    \subfloat[]{\includegraphics[width=0.245\linewidth]{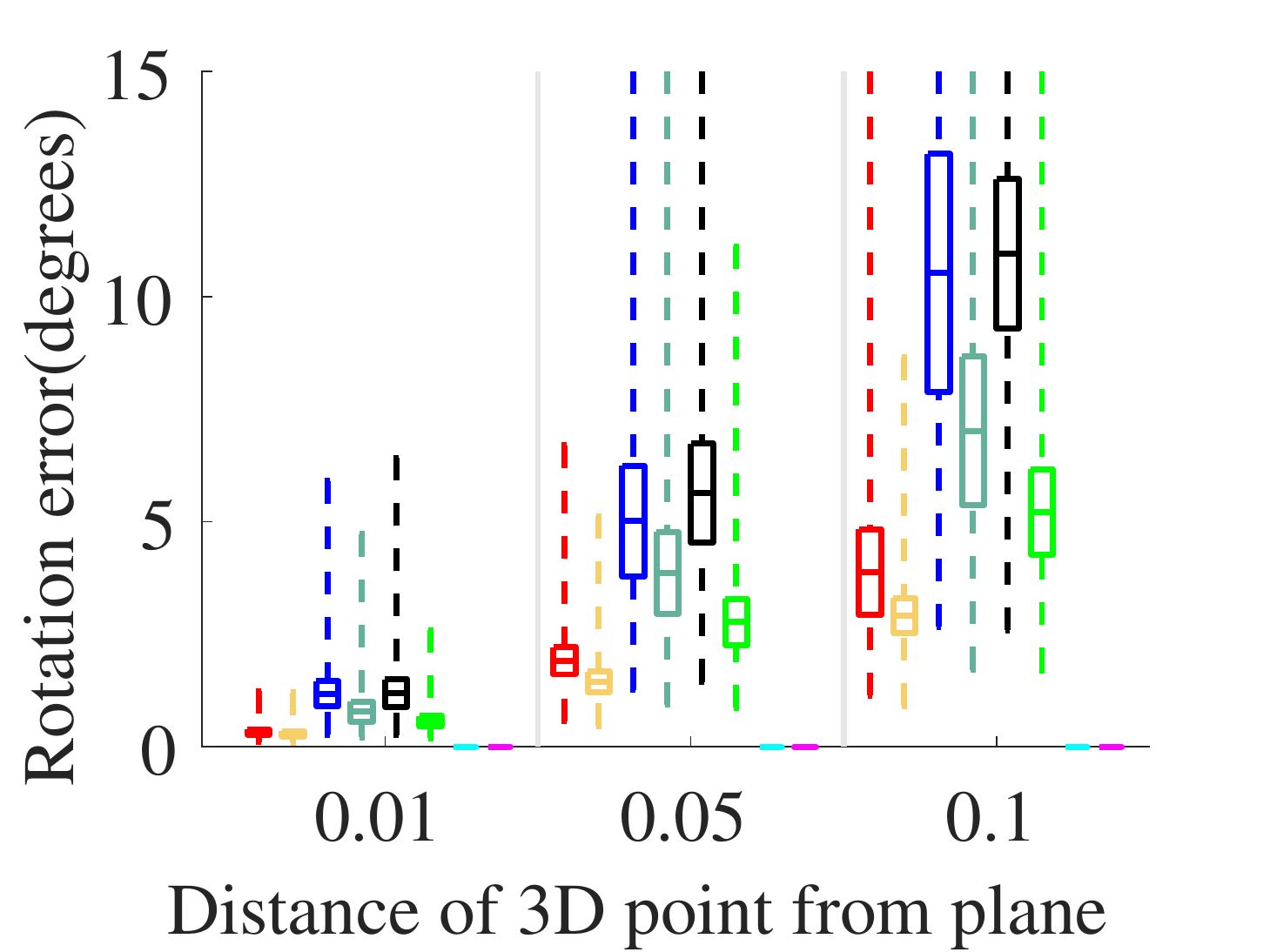}}
     \subfloat[]{\includegraphics[width=0.245\linewidth]{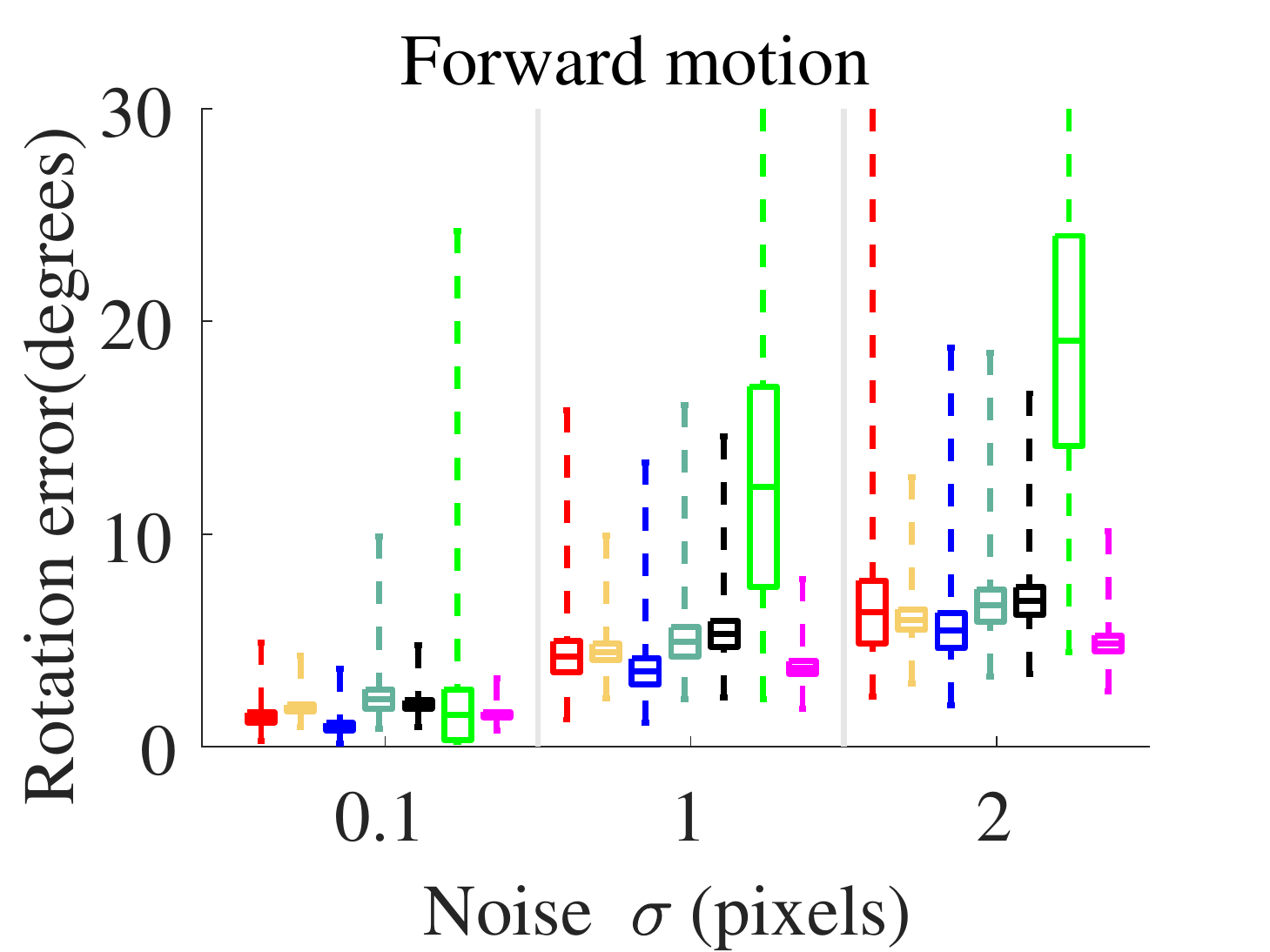}}\\
    \subfloat[]{\includegraphics[width=0.245\linewidth]{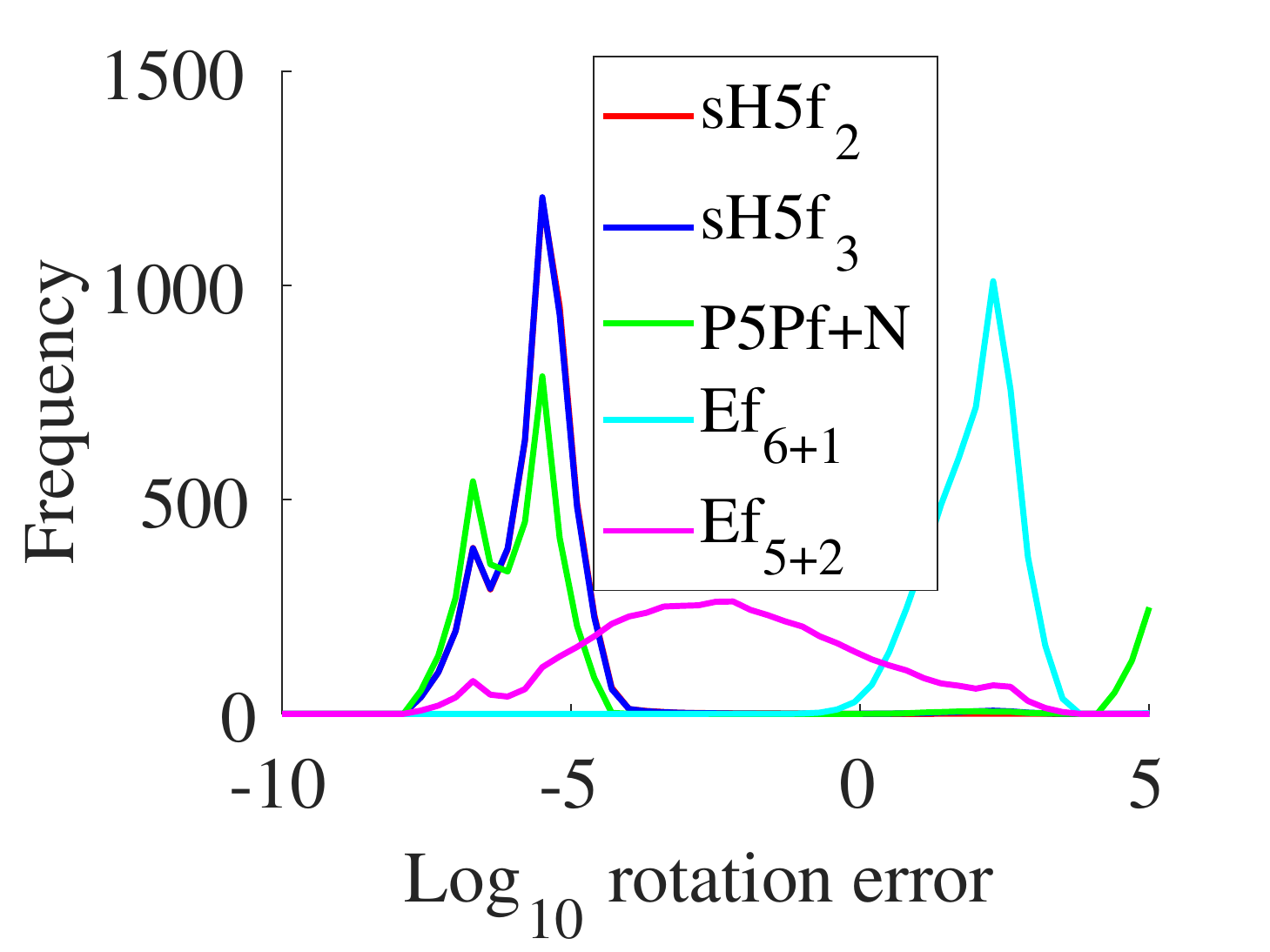}}
    \subfloat[]{\includegraphics[width=0.245\linewidth]{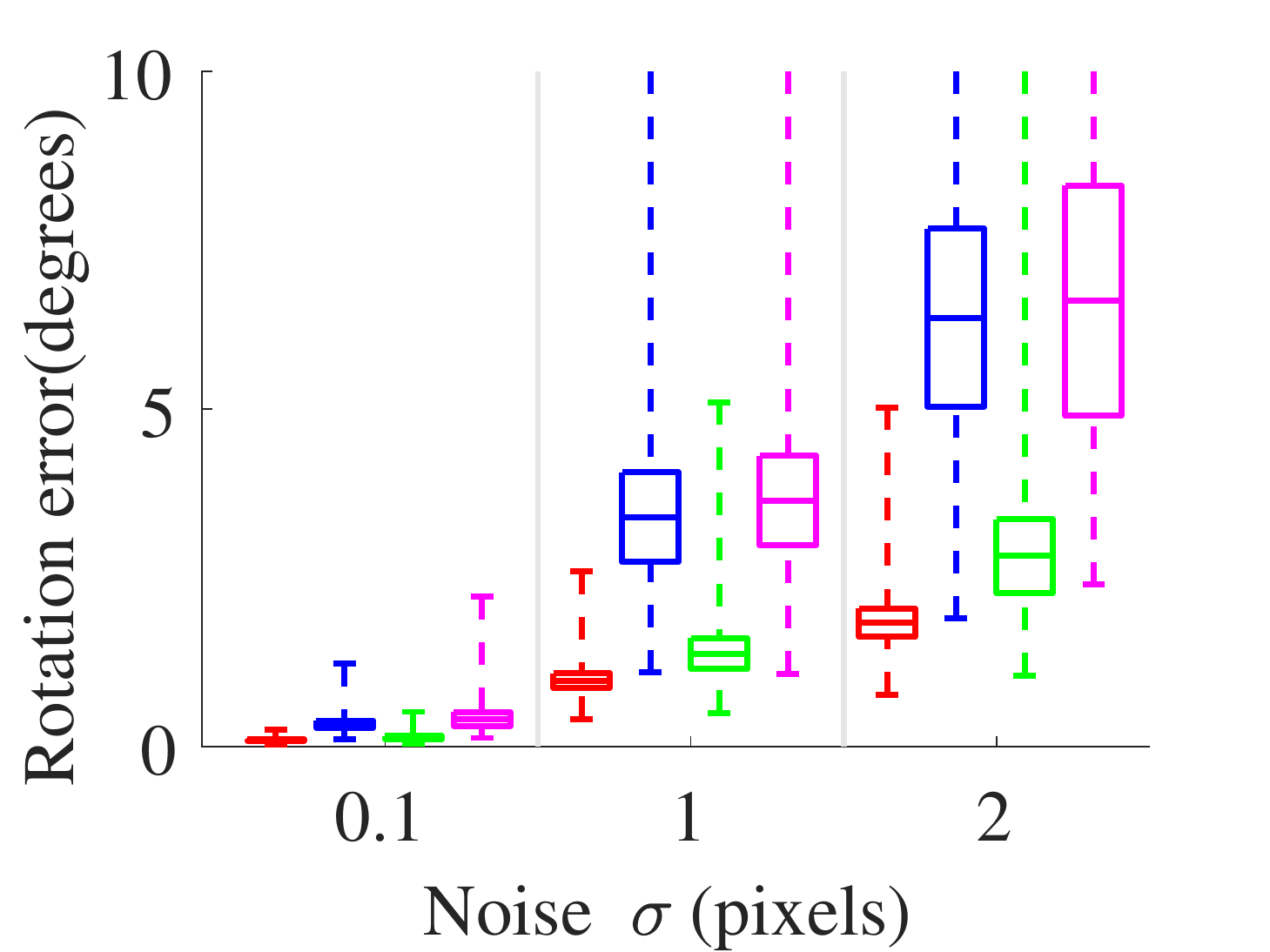}}
    \subfloat[]{\includegraphics[width=0.245\linewidth]{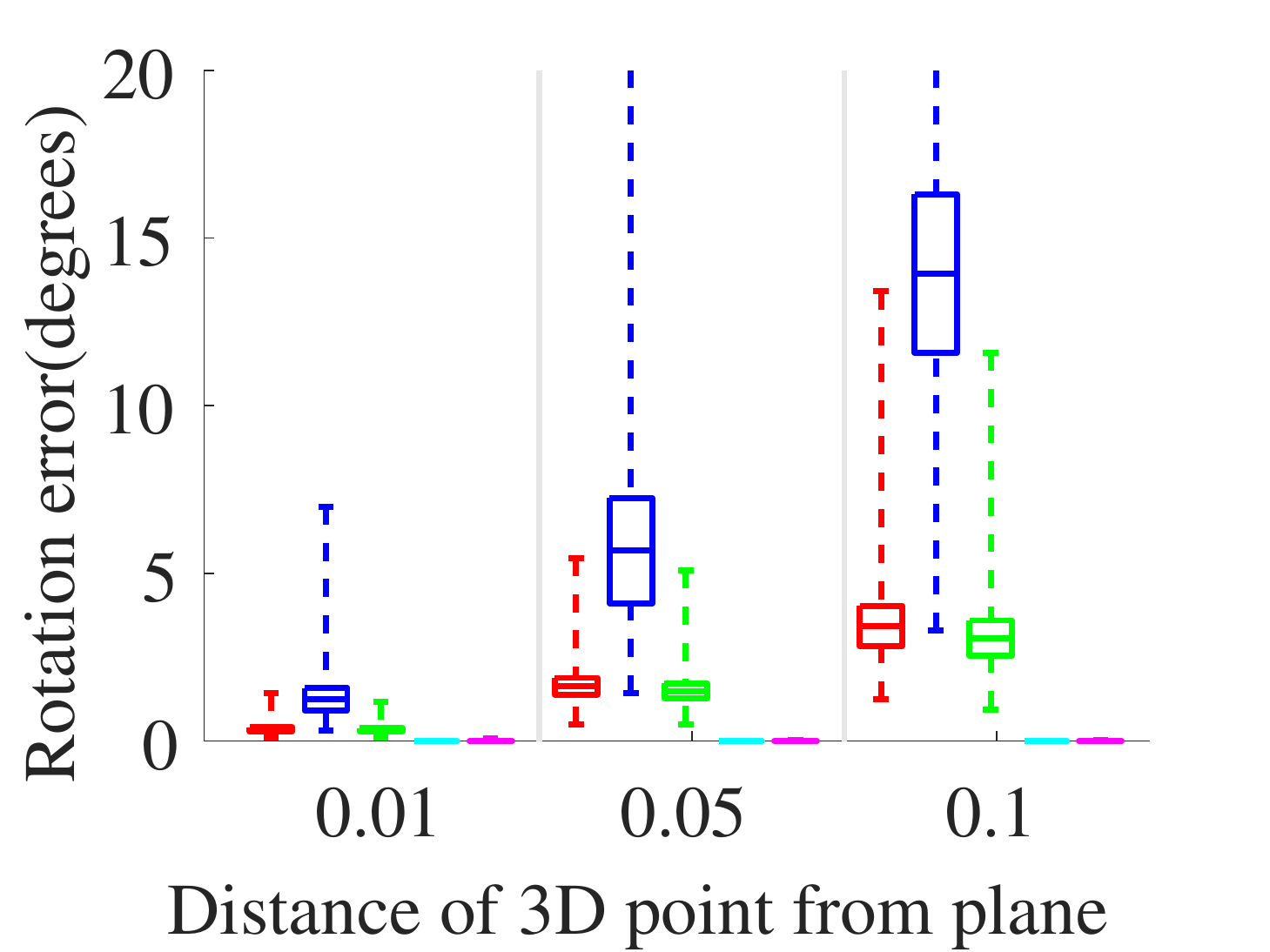}}
     \subfloat[]{\includegraphics[width=0.245\linewidth]{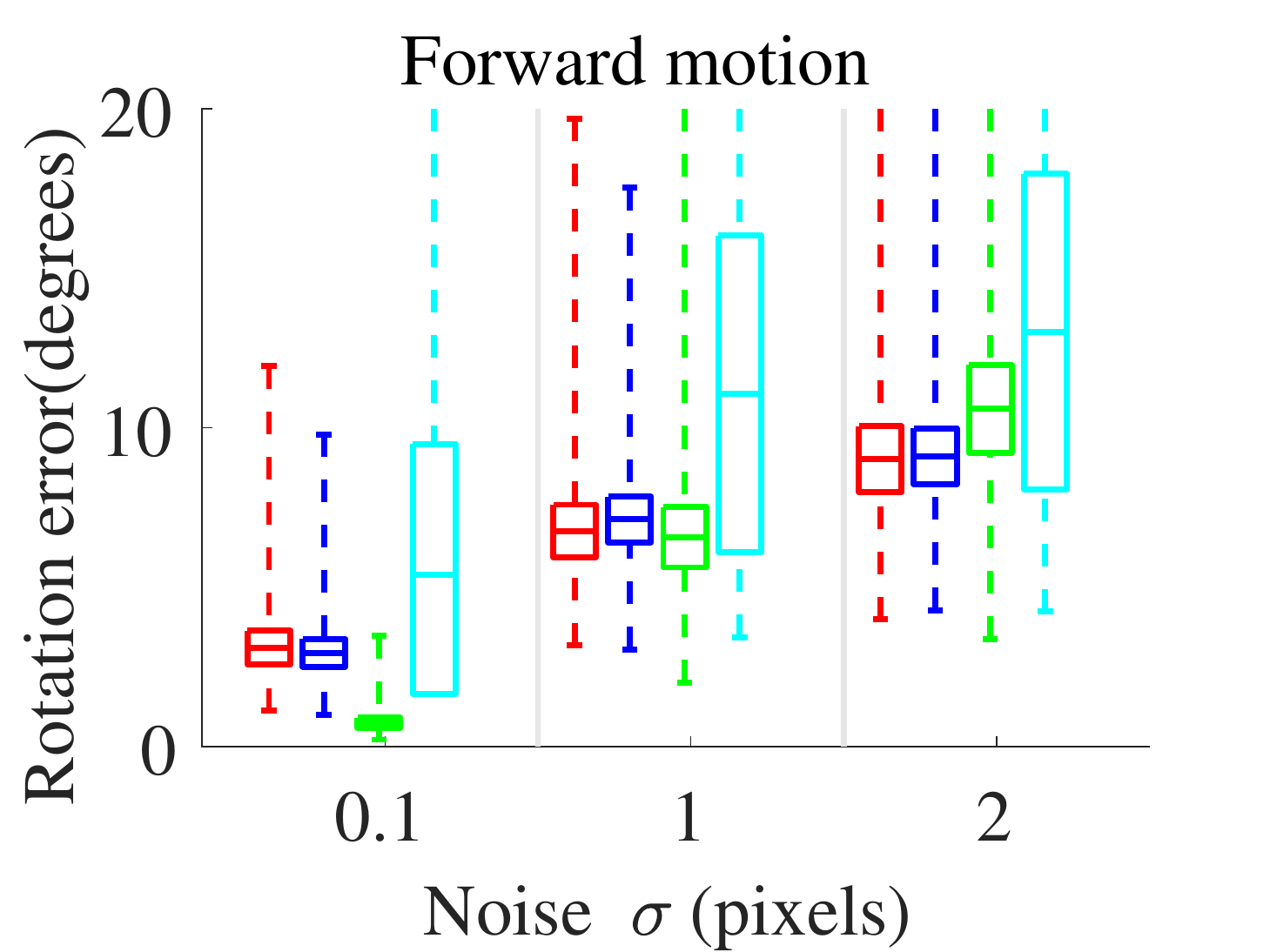}}
\caption{ \textbf{Top}: solvers for calibrated cameras. \textbf{Bottom}: solvers for partially calibrated cameras: (a,e) numerical stability, (b,f) performance in the presence of image noise, (c,g) close-to-planar scenes, (d,h) forward motion in the presence of image noise.
% (d,h) Forward motion in the presence of image noise.  
%\textbf{Bottom}: Solvers for partially calibrated cameras: (e) Numerical stability, (f) Performance in the presence of image noise. (g) Close-to-planar scenes. (h) Forward motion in the presence of image noise.  
% (a) A study of numerical stability of the solver. Ground truth error in $\M{R}$ for cameras with unknown focal length in the presence of (b) image noise, (c) Non-planar scene and (d) Forward camera motion
}
\label{fig:synthgraphs}
\end{figure*}

\noindent
% In this section, we study 
This section studies the performance of the proposed solvers, $\mathbf{s\M{H}5_2}$, $\mathbf{s\M{H}5_3}$, $\mathbf{s\M{H}4.5_2}$, $\mathbf{s\M{H}4.5_3}$, $\mathbf{s\M{H}5_4}$, $\mathbf{s\M{H}5f_2}$, and $\mathbf{s\M{H}5f_3}$, both on synthetic and real-world images. 
For comparison, we use four state-of-the-art minimal solvers for estimating the semi-generalized epipolar geometry~\cite{Wu_iccv15}, \ie, the $\mathbf{\M E_{5+1}}$, $\mathbf{\M E_{4+2}}$, $\mathbf{\M Ef_{6+1}}$, and $\mathbf{\M Ef_{5+2}}$ solvers. 
Note, that for the experiments where we do not need the scale of the translation, $\mathbf{\M E_{5+1}}$ reduces to the well-known 5pt solver $\mathbf{\M E5}$~\cite{nister2004efficient} and the $\mathbf{\M Ef_{6+1}}$ problem reduces to the one-sided focal length 6pt solver $\mathbf{\M E6f}$~\cite{bujnak20093d}. In such  experiments we also consider the 4pt homography solver $\mathbf{\M H4}$~\cite{hartley2003multiple}. We excluded the $\mathbf{\M Ef_{5+2}}$ solver from real experiments since it was too slow when used on large datasets inside RANSAC~\cite{fischler1981random}.
For a fair comparison, we further consider $\mathbf{\text{P3P/P5Pf}+\V N}$ solvers 
% We designed these solvers 
designed for %this particular 
our semi-generalized homography setup:  %In these solvers, 
we first use three correspondences between two calibrated generalized cameras $\cam{G}_i$ and $\cam{G}_j$ to estimate the normal $\V n$ and intercept $d$ of the observed plane $\pi$. 
% This information 
The plane is then used to lift %transform 
3/5 2D-2D %correspondences 
matches between the perspective camera $\cam{P}$ and arbitrary cameras in the generalized camera $\cam{G}$ to 2D-3D correspondences. 
Finally, the pose of the perspective camera is computed using the %well-known 
P3P~\cite{Persson_2018_ECCV} or P5Pf~\cite{KukelovaP5pfr} solvers\footnote{
In the unknown focal length case, we use the non-minimal P5Pf solver since the available implementations of the minimal P4Pf one were either much slower~\cite{Bujnak08CVPR} than the P5Pf solver or did not work for planar scenes~\cite{kukelova2016efficient}.}.  Note that these $\text{P3P/P5Pf}+\V N$ solvers, compared to the proposed solvers, require point %correspondences 
matches between two cameras $\cam{G}_i$ and $\cam{G}_j$ in the generalized camera. 
However, these correspondences are used only in the first step, when estimating the normal and plane intercept. 
% Therefore, 
Thus, they do not need to be visible in %camera 
$\cam{P}$ as in the standard 2D-3D pipeline. 
We are not aware that such solvers have been used in the literature. 
% To the best of our knowledge such solvers have not been considered in the literature before. 

\subsection{Synthetic Scenes} \label{sec:synth}
\noindent
% We studied 
We study the performance of our proposed solvers on synthetically generated 3D scenes with known ground truth. % parameters. 
The 3D points are randomly distributed on a plane of size $10 \times 10$. 
Each 3D point is projected into up to six cameras with realistic focal lengths. 
Five of these cameras represent the generalized camera $\cam{G}$
%, respectively, for the calibrated or unknown focal length solvers, 
and one camera is considered as the camera $\cam{P}$. 
The orientations and positions of the cameras are selected at random such that they roughly look towards the scene from a random distance, varying from $20$ to $35$, from the plane. 
The simulated images have a resolution of $1000 \times 1000$~px. 
Here, we focus on the errors in the estimated rotations $\M{R}$ for the calibrated and %solvers as well as the 
unknown focal length solvers. 
% and the errors in the focal length $f$ for 
The rotation error is computed as the angle in the axis-angle representation of $\M{R}_{GT}^{-1} \M{R}$, where $\M{R}_{GT}$ is the ground truth and $\M{R}$ is the estimated rotation. 
Plots for the position and focal length errors can be found in the SM. %are, due to the lack of space, in the SM. %supplementary material.
% We next compare the numerical stability of our proposed solvers as well their performance in the presence of image noise and close-to-planar scenes with respect to that of the competitive solvers. Additionally we have also evaluated the performance of the competitive solvers for a forward moving camera in the presence of image noise, which is described in the supplementary material. 

% \vspace{-10pt}
\PAR{Numerical stability.} \label{sec:nstab}
% In order to 
We measure the numerical stability of the solvers 
%proposed solvers 
by evaluating 5k camera setups for planar scenes. 
We compare the accuracy of the rotations estimated by the proposed solvers $\mathbf{s\M{H}5_2}$, $\mathbf{s\M{H}4.5_2}$, $\mathbf{s\M{H}5_3}$, $\mathbf{s\M{H}4.5_3}$, and $\mathbf{s\M{H}5_4}$ with that of $\mathbf{\M E_{5+1}}$, $\mathbf{\M E_{4+2}}$, and $\text{P3P}+\V N$ in Fig.~\ref{fig:synthgraphs}(a). 
% While o
Our %proposed 
solvers, $\mathbf{s\M{H}5_2}$, $\mathbf{s\M{H}5_3}$, and $\mathbf{s\M{H}5_4}$, achieve better stability with fewer failures (\ie, no peak on the right side). 
$\mathbf{s\M{H}4.5_2}$ and $\mathbf{s\M{H}4.5_3}$ have comparable stability as %compared to 
the other solvers.
% Similarly, 
Fig.~\ref{fig:synthgraphs}(e) compares %provides a comparison of 
the numerical stability of 
%the proposed 
$\mathbf{s\M{H}5f_2}$ and $\mathbf{s\M{H}5f_3}$ with that of the solvers $\mathbf{\M Ef_{6+1}}$, $\mathbf{\M Ef_{5+2}}$, and $\mathbf{\text{P5Pf}+\V N}$. 
%in terms of relative rotation error. 
%The proposed solvers achieve state-of-the-art stability.
%We also note that the stability experiments being conducted on planar scenes created a degenerate configuration for the $\mathbf{\M Ef_{6+1}}$ solver which
%explains the reported performance. 
Note that a planar scene is a degenerate configuration for the $\mathbf{\M Ef_{6+1}}$ solver, which
explains the reported performance.

% \vspace{-8pt}
\PAR{Image noise.} \label{sec:noise}
Next, we test the performance of all solvers in the presence of Gaussian noise with standard deviation $\sigma$, added to the image points in all cameras. Fig. \ref{fig:synthgraphs}(b,f) show the  rotation error (in degrees) for solvers for calibrated (b) as well as partially calibrated (f) cameras.
Here, we depict the results as box plots %using the {\sc Matlab} function {\tt boxplot} 
which show the $\textrm{25}\%$ to $\textrm{75}\%$ quantile values as boxes with a horizontal line for the median. We note that our proposed solvers $\mathbf{s\M{H}5_2}$, $\mathbf{s\M{H}4.5_2}$, and $\mathbf{s\M{H}5f_2}$ have better or comparable performance than the  competing %competitive 
solvers in the presence of image noise. Moreover, we observe that $\mathbf{s\M{H}4.5_2}$ and $\mathbf{s\M{H}4.5_3}$ 
are more stable than  
% have better stability than that of
$\mathbf{s\M{H}5_2}$ and $\mathbf{s\M{H}5_3}$, respectively, in the presence of image noise.

% \vspace{-8pt}
\PAR{Close-to-planar scenes.} \label{sec:closetoplanar} 
We also consider the case where the scene is close to being entirely planar by placing the scene plane at $z=0$ and sampling 3D points with varying plane-to-point distances. 
Fig.~\ref{fig:synthgraphs}(c,g) show the rotation errors %in the estimated rotations 
for %solvers assuming 
calibrated (c) and partially calibrated (g) cameras. 
% The proposed solvers 
Our $\mathbf{s\M{H}5_2}$ and $\mathbf{s\M{H}4.5_2}$ solvers are more accurate than $\mathbf{\text{P3P}+\V N}$ while $\mathbf{s\M{H}5f_2}$ has comparable stability to $\mathbf{\text{P5Pf}+\V N}$ for close-to-planar scenes. 
As expected, %It is clear that 
the accuracy %quality of the estimates obtained using the 
of the proposed solvers deteriorates with the increasing non-planarity of the scene. However, the errors, even for larger non-planarity, are comparable %correspond 
to the errors obtained by all %competing 
general solvers in the presence of $2$ px image noise.

% \vspace{-8pt}
\PAR{Forward motion with image noise.} \label{sec:forwardmotion}
Figures \ref{fig:synthgraphs}(d,h) show the %ground-truth 
rotation error for %in the estimated rotation by 
% the solvers for 
calibrated (d) and %as well as 
partially calibrated (h) cameras. Our %proposed 
solvers for calibrated cameras have similar or better stability than the competing solvers. For the unknown focal length case, %solvers
%, we note that 
our proposed solvers lead 
%to reasonable rotation estimates. 
to similar rotation estimates than the competing solvers.
% to rotation. 
We note that in case of a pure forward motion, the solver $\mathbf{\M Ef_{5+2}}$ either failed or led to very unstable results. As a result of this, we have not considered the solver $\mathbf{\M Ef_{5+2}}$ in the graphs.

\subsection{Computational Complexity}
\noindent Tab.~\ref{table:computationalcomplexity} reports the computational complexity of the studied solvers. Since we do not have equally efficient C++ implementations of all solvers (some solvers are highly optimized, \eg, $\mathbf{\M E5}$ and $\mathbf{\text{P3P}}$  from the PoseLib~\cite{PoseLib} library, while some do not contain any special optimization, \eg, $\mathbf{\M{E}_{4+2}}$, $\mathbf{\M{E}f_{5+2}}$~\cite{Wu_iccv15}), we compare only the most time consuming operations performed by these solvers. 
% For this w
We thus %consider 
focus on the matrix size for each %of the 
critical matrix operation.
%, \eg, $4 \times 8$ for G-J operation denotes a Gauss-Jordan elimination of a matrix of size $4\times 8$. 

\subsection{Real-World Experiments}
\noindent %In the proposed solvers, we 
Our solvers built on the fact that man-made environments frequently contain planes and planar structures both indoors and outdoors. 
% However, to demonstrate their applicability in practice, we tested the solvers on general real-world data. 
% Our objective is t
To show the usefulness of our solvers in real applications like visual odometry and visual localization, 
we test them on general real-world data. 
Such general scenarios will of course give 
% marginal 
an advantage to our competitors, \eg,  \cite{nister2004efficient}, that work for planar as well as non-planar scenes. 
Yet, we show that our new proposed solvers return comparable pose and focal length estimates and sometimes even outperform the state-of-the-art general solvers. % for estimating the epipolar geometry or absolute pose. 
As such, we believe that our solvers can be combined with existing ones in a hybrid RANSAC~\cite{camposeco2018hybrid}, where the most suitable solver is selected for each scene in a data-dependent manner.

\begin{figure*}[t]
    \centering
    \begin{tabular}[b]{c}
        \includegraphics[trim={1mm 0mm 10mm 0mm},clip, width=0.31\textwidth]{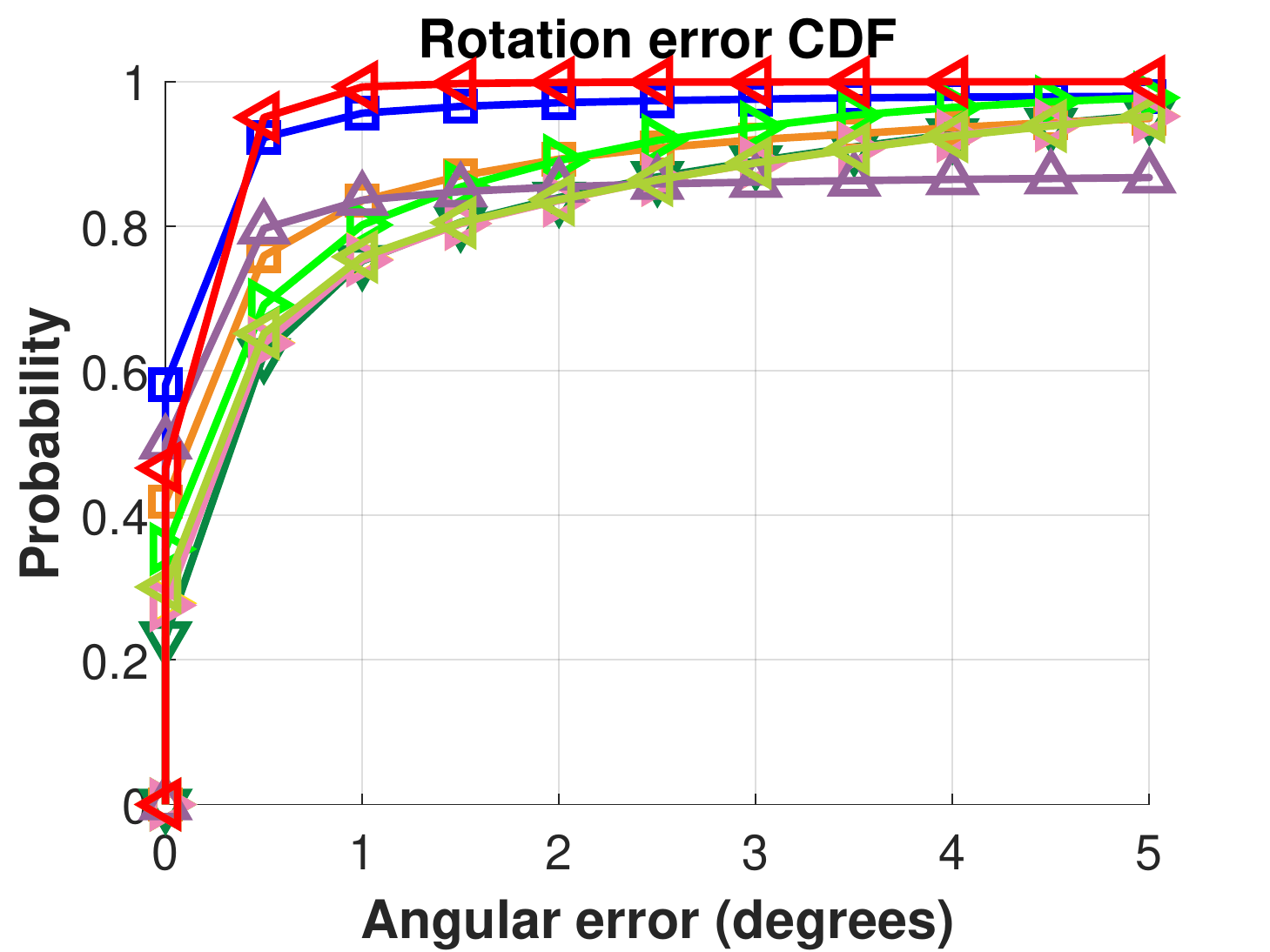}
        \includegraphics[trim={1mm 0mm 10mm 0mm},clip, width=0.31\textwidth]{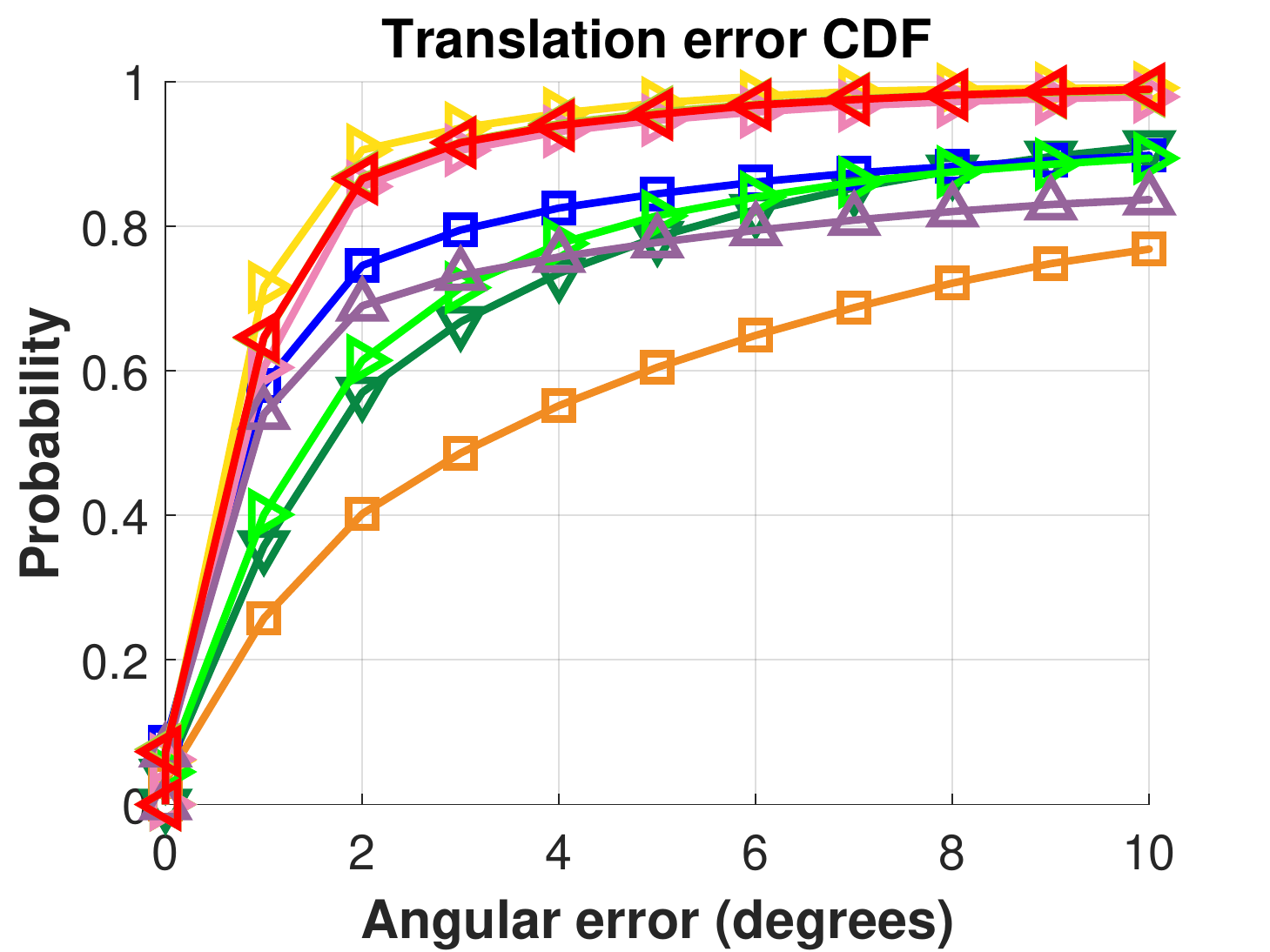}
        \includegraphics[trim={1mm 0mm 10mm 0mm},clip, width=0.31\textwidth]{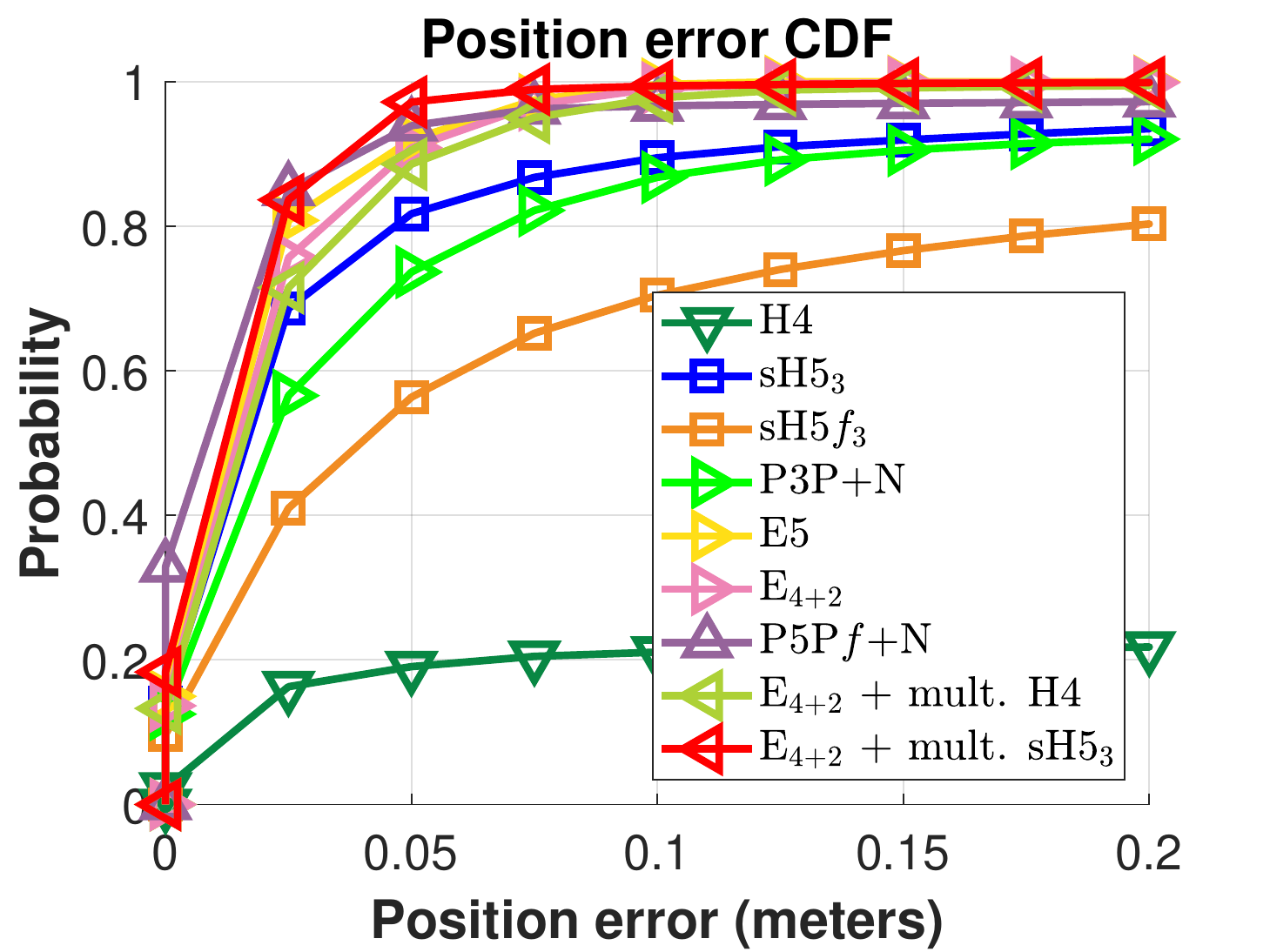}
    \end{tabular}
    % \vspace{-12pt}
    \caption{ The CDFs of the rotation, translation (degrees) and position (meters) errors on $23,190$ image pairs from the KITTI dataset. 
    More accurate methods are closer to the top-left corner. 
    Since most tested methods do not return the translation scale due to estimating the relative pose, we used the scale from the ground truth path to calculate the position error. Tab.~\ref{tab:kitti_results} shows the corresponding error values. }
    \label{fig:kitti_experiments}
    % \vspace{-6pt}
\end{figure*}

% \vspace{-8pt}
\PAR{Localization experiment.}
We evaluate %the performance of 
all variants of our $\mathbf{s\M{H}5}$ and  $\mathbf{s\M{H}4.5}$ solvers for calibrated cameras in the context of visual localization. 
We use the %King's College, Old Hospital, Shop Facade, and St. Mary's Church 
subset of scenes from the Cambridge Landmarks dataset~\cite{Kendall2015ICCV} commonly used in the literature~\cite{Sattler2019CVPR}. 
Note that while these scenes contain one or more dominant planes, none of them is perfectly planar. 
% More localization experiments can be found in the SM.
% We report the median position and orientation error for each method. 

% \begin{table}[t!]
% \begin{center}
% \setlength{\tabcolsep}{1.5pt}
% \footnotesize{
% \begin{tabular}{|l|c|c|c|c|c|}
% \hline
%  Method & King's & Old & Shop & St. Mary \\ \hline
% $\mathbf{s\M{H}5_3}$ (top-2) (ours) & 0.58\,/\,0.76 & 2.16\,/\,3.88 & 0.14\,/\,0.60 & 1.19\,/\,3.46 \\ \hline
% $\M{E}_{4+2}$~\cite{Zheng2015ICCV} (top-2) & 1.25\,/\,1.91 & 2.33\,/\,3.72 & 0.19\,/\,0.79 & 1.07\,/\,2.45 \\ \hline
% $\M{E}_{5+1}$~\cite{Zheng2015ICCV} (top-2) & 0.41\,/\,0.59 & 1.07\,/\,2.28 & 0.11\,/\,0.42 & \textbf{0.27}\,/\,\textbf{0.80} \\ \hline \hline 
% $\mathbf{s\M{H}5_2}$ \& $\mathbf{s\M{H}5_3}$ (top-10) & \textbf{0.24}\,/\,\textbf{0.42} & \textbf{0.85}\,/3.29 & \textbf{0.07}\,/\,\textbf{0.31} & 0.44\,/\,1.67 \\ \hline
% Sift+5Pt~\cite{Zhou2020ICRA} & 0.48\,/\,1.13 & 0.88\,/\,\textbf{1.91} & 0.17\,/\,0.99 & 0.35\,/\,1.58 \\ \hline
% \end{tabular}
% }
% \end{center}
% \caption{Localization results on the Cambridge Landmarks dataset~\cite{Kendall2015ICCV}. We report the median position (in meters) and rotation (in degrees) errors. For Sift+5Pt, we report the detailed results from~\cite{zhou2020learn} instead of the aggregated results from~\cite{Zhou2020ICRA}.}%
% \label{tab:loc_results_old}%
% \end{table}

\begin{table*}[t!]
\begin{center}
\setlength{\tabcolsep}{1.5pt}
\footnotesize{
\begin{tabular}{|l|c|c|c|c|c|c|c|c|c|c|c|c|c|c|c||c|c|}
\hline
  & \multicolumn{3}{c|}{King's College} & \multicolumn{3}{c|}{Old Hospital} & \multicolumn{3}{c|}{Shop Facade} & \multicolumn{3}{c|}{St. Mary Church} & \multicolumn{2}{c|}{Avrg. all} \\ \cline{2-15}
 Method & pos. & rot. & time & pos. & rot. & time & pos. & rot. & time & pos. & rot. & time & pos. & rot. \\ \hline

 $\mathbf{\M{E}_{5+1}}$~\cite{Wu_iccv15}   (100 iter.) & 
 \best{0.20}\,/\,\secondBest{0.44} & \secondBest{0.36}\,/\,\secondBest{0.61} & 0.29 & 
 \secondBest{0.54}\,/\,\best{1.30} & \best{1.02}\,/\,\best{2.12} & 0.15 & 
 \best{0.06}\,/\,\best{0.10} & \secondBest{0.33}\,/\,\best{0.46} & 0.15 & 
 \best{0.13}\,/\,\best{0.20} & \best{0.51}\,/\,\best{0.73} & 0.18 & \best{0.23} & \best{0.56}\\ \hline
$\M{E}_{4+2}$~\cite{Wu_iccv15} (100 iter.) & 
 0.25\,/\,1.58 & 0.42\,/\,1.70 & 0.26 & 
 1.51\,/\,56.2 & 2.82\,/\,6.90 & 0.15 & 
 0.09\,/\,2.77 & 0.44\,/\,3.28 & 0.14 & 
 0.41\,/\,242.4 & 1.42\,/\,5.62 & 0.17 & 0.57 & 1.28\\ \hline
\rowcolor{mygray} \textbf{ours} ($\mathbf{s\M{H}5}$)  (100 iter.) & 
 \secondBest{0.22}\,/\,0.71 & 0.39\,/\,1.20 & \best{0.20} & 
 0.88\,/\,\secondBest{2.20} & \secondBest{1.68}\,/\,3.98 & \best{0.08} & 
 0.09\,/\,0.78 & 0.43\,/\,2.23 & \best{0.08} & 
 0.25\,/\,2.52 & 0.95\,/\,6.50 & \best{0.11} & \secondBest{0.36} & \secondBest{0.86}\\ \hline
\rowcolor{mygray} \textbf{ours} ($\mathbf{s\M{H}4.5}$)  (100 iter.) & 
 \best{0.20}\,/\,\best{0.32} & \best{0.33}\,/\,\best{0.49} & \secondBest{0.23} & 
 \best{0.51}\,/\,48.4 & \best{1.02}\,/\,\secondBest{3.15} & \secondBest{0.12} & 
 \secondBest{0.07}\,/\,\secondBest{0.14} & \best{0.32}\,/\,\secondBest{0.67} & \secondBest{0.11} & 
 \secondBest{0.15}\,/\,\secondBest{0.30} & \secondBest{0.52}\,/\,\secondBest{1.24} & \secondBest{0.14} & \best{0.23} & \best{0.55}\\ \hline \hline
 
%% 1000 iterations
$\mathbf{\M{E}_{5+1}}$~\cite{Wu_iccv15}   (1k iter.) & 
 \best{0.19}\,/\,0.33 & \secondBest{0.34}\,/\,\secondBest{0.48} & 0.82 & 
 \best{0.42}\,/\,\secondBest{1.10} & \secondBest{0.75}\,/\,\best{1.78} & 0.39 & 
 \best{0.06}\,/\,\best{0.10} & \best{0.29}\,/\,\best{0.44} & 0.37 & 
 \best{0.11}\,/\,\best{0.15} & \best{0.38}\,/\,\best{0.55} & 0.49 & \best{0.20} & \best{0.44}\\ \hline
$\mathbf{\M{E}_{4+2}}$~\cite{Wu_iccv15} (1k iter.) & 
 \secondBest{0.20}\,/\,0.42 & 0.35\,/\,0.60 & 0.75 & 
 0.83\,/\,2.51 & 1.55\,/\,3.88 & 0.56 & 
 \secondBest{0.07}\,/\,\secondBest{0.16} & \secondBest{0.32}\,/\,0.70 & 0.53 & 
 0.20\,/\,0.70 & 0.71\,/\,2.17 & 0.59 & 0.33 & 0.73 \\ \hline 
\rowcolor{mygray} \textbf{ours} ($\mathbf{s\M{H}5}$)  (1k iter.) & 
 \secondBest{0.20}\,/\,\secondBest{0.31} & \secondBest{0.34}\,/\,\secondBest{0.48} & \best{0.33} & 
 0.46\,/\,\best{1.03} & 0.89\,/\,2.47 & \best{0.16} & 
 \best{0.06}\,/\,\best{0.10} & \best{0.29}\,/\,\secondBest{0.45} & \best{0.16} & 
 0.13\,/\,0.43 & 0.47\,/\,1.35 & \best{0.20} & \secondBest{0.21} & \secondBest{0.50}\\ \hline
\rowcolor{mygray} \textbf{ours} ($\mathbf{s\M{H}4.5}$)  (1k iter.) & 
 \best{0.19}\,/\,\best{0.30} & \best{0.33}\,/\,\best{0.46} & \secondBest{0.52} & 
 \best{0.40}\,/\,1.21 & \best{0.74}\,/\,\secondBest{1.91} & \secondBest{0.27} & 
 \best{0.06}\,/\,\best{0.10} & \best{0.29}\,/\,\best{0.44} & \secondBest{0.26} & 
 \secondBest{0.12}\,/\,\secondBest{0.17} & \secondBest{0.40}\,/\,\secondBest{0.59} & \secondBest{0.33} & \best{0.20} & \best{0.44} \\ \hline \hline

Sift+5Pt~\cite{Zhou2020ICRA,zhou2020learn} & 0.48 / - & 1.13 / - & - &  0.88 / - & {1.91} / - & - & 0.17 / -  & 0.99 / - & - & 0.35 / - & 1.58 / - & - & 0.47 & 0.88\\ \hline
\end{tabular}
}
\end{center}
\vspace{-6pt}
\caption{Localization results on Cambridge Landmarks~\cite{Kendall2015ICCV}. We report the median/mean position (in meters) and rotation (in degrees) errors, and the mean RANSAC time (in seconds). We also report the average median position and rotation error over all four scenes. We show results for fixing the number of RANSAC iterations to 100 respectively 1000. Best and second best results are shown in red and blue.}% \todo{Recalculate average of medians.}}% RANSAC iterations. }%For Sift+5Pt~\cite{Zhou2020ICRA}, we report the detailed results from~\cite{zhou2020learn}.}% instead of the aggregated results from~\cite{Zhou2020ICRA}.}%
\label{tab:loc_results}%
\end{table*}

Our $\mathbf{s\M{H}5}$ and $\mathbf{s\M{H}4.5}$ and the %solvers, as well as the solvers 
$\mathbf{\M{E}_{4+2}}$, and $\mathbf{\M{E}_{5+1}}$ solvers enable a particularly 
% a 
light-weight type of structure-less localization pipelines that do not need to store a 3D model. 
% The resulting %in a 
Such representations %that 
can be easily maintained~\cite{Torii2019TPAMI}.  
In contrast to $\mathbf{\text{P3P}+\V N}$ and %the 
SfM-on-the-fly% approach from
~\cite{Torii2019TPAMI}, %they only require 
our solvers only need matches between the pinhole image and the generalized camera images but not within images in the generalized camera. % are needed. 
This %, thus 
keeps feature matching to a minimum. 
We implement such a pipeline by using DenseVLAD-based image retrieval~\cite{Torii2015CVPR} to identify the %top-
$10$ reference images most similar to a given query. 
The generalized camera is then defined using the known poses of the retrieved images. 

% Our solvers can be easily integrated into RANSAC with local optimization
We integrate our solvers into RANSAC with local optimization (LO-RANSAC)~\cite{Lebeda2012BMVC,Sattler2019Github}. 
% Given a test image, RANSAC simply randomly samples 5 matches among the matches found with the retrieved images. 
In each iteration, we simply randomly sample 5 matches from all matches found with the retrieved images. 
We then select the most suitable solver for this sample, \eg, $\mathbf{s\M{H}5_4}$ if four matches come from the same reference image. % or $\mathbf{s\M{H}4.5_2}$ if at most two matches come from the same reference image.  
This approach is possible thanks to the fact that our solvers cover  all possible combinations of 5 point correspondences.
However, this approach is not suitable for the 
%The 
$\mathbf{\M{E}_{4+2}}$ and $\mathbf{\M{E}_{5+1}}$ solvers as the chance of randomly sampling 4 or more matches from the same reference image is very small. %\footnote{Out of 1.158M RANSAC samples drawn for our $\mathbf{s\M{H}4.5}$ solvers, less than 0.7\% of the samples had 4 matches from the same reference image.}. % Due to the rareness of this event, we actually skip these samples rather calling the $\mathbf{s\M{H}4.5}_4$ solver.} 
Instead, we first randomly select two ($\mathbf{\M{E}_{5+1}}$) or three ($\mathbf{\M{E}_{4+2}}$)\footnote{Note that two of the three images can be identical.} retrieved reference images. % uniformly at random and then randomly select the required matches from the selected images. 
We then randomly select the required matches from these images. 
This sampling scheme is incompatible with RANSAC's 
% As a result, RANSAC's 
standard stopping criterion. % is not directly applicable. 
For a fair comparison, we thus run LO-RANSAC for each solver for a fixed number of iterations. 
The best model found by LO-RANSAC is refined over all inliers (see SM for details).
% For all solvers, the best model returned by RANSAC is refined over all inliers (see SM for details).

% require matches to come from two cameras in a generalized camera. 
% Thus, we use only the top-2 retrieved images. 
% For a fair comparison, we compare against using $\mathbf{s\M{H}5_3}$ for the top-2 retrieved images. 
% Both $\mathbf{s\M{H}5_3}$ and $\mathbf{s\M{H}5_2}$ can handle more images and we use the top-10 retrieved images to demonstrate the advantages of this property. 
% In this case, we adaptively choose between $\mathbf{s\M{H}5_2}$ and $\mathbf{s\M{H}5_3}$ inside RANSAC, selecting the solver that is applicable for a given sample. 
% All methods are integrated into LO-RANSAC~\cite{lebeda2012fixing,Sattler2019Github}. 
% See the supplementary material for details. 
% % and local optimization is performed by non-linear optimization of reprojection errors (either obtained via the semi-generalized homography or by triangulating points based on the estimated pose), which is implemented in Ceres~\cite{ceres-solver}. 

% Tab.~\ref{tab:loc_results} shows the results of our experiments. 
As shown in Tab.~\ref{tab:loc_results}, our solvers outperform the $\mathbf{\M{E}_{4+2}}$ solver. 
They are consistently among the top-2 approaches based on mean / median position and orientation errors and lead to the fastest RANSAC times. %but not the $\M{E}_{5+1}$ solver. 
% However, using more images to define the generalized cameras, as afforded by the $\mathbf{s\M{H}5_2}$ solver, can lead to a significant improvement in most scenes. 
% The fact that $\M{E}_{5+1}$ still performs better on St. Mary's Church can be attributed to the fact that this scene contains multiple parts without large planar structures. 
Averaged over all datasets, our $\mathbf{s\M{H}4.5}$ solvers lead to the same median 
% and better average 
results as the $\mathbf{\M{E}_{5+1}}$ solver at faster run-times. 
The results clearly show the usefulness of our solvers. % in practice. 
In particular, our results point towards an interesting %direction for future research: 
research direction: 
our faster solvers can be used to quickly estimate %get an estimate for 
the inlier ratios for each reference image. 
This can then be used for guided sampling of image pairs for the $\mathbf{\M{E}_{5+1}}$ solver, \eg, inside a hybrid RANSAC scheme\footnote{%Note that t
The hybrid RANSAC formulation from~\cite{camposeco2018hybrid} deals with two sources of matches and cannot be easily extended to %adapted to handle 
more sources (each retrieved reference images represents a source with its own inlier ratio).}. % As such, we did not use hybrid RANSAC in our experiments.}.  
This approach should deliver the best from both types of solvers. 

% For comparison, 
Tab.~\ref{tab:loc_results} also includes the Sift+5pt approach~\cite{Zhou2020ICRA,zhou2020learn}, which estimates the relative pose between the query and retrieved images based on SIFT feature matches and essential matrix estimation. The relative poses and the known absolute poses of the retrieved images %, with unknown scale of translation, 
are then used to estimate the query pose. %triangulate the query pose based on the known poses of the retrieved images. 
Our approach consistently outperforms~\cite{Zhou2020ICRA,zhou2020learn}. %by a wide margin. 

% , $\mathbf{s\M{H}5f_2}$ and $\mathbf{s\M{H}5f_3}$ solvers on both synthetic and real images. For comparison, we use four state-of-the-art minimal solvers for the semi-generalized epipolar geometry~\cite{Wu_iccv15}, \ie, the $\mathbf{\M E_{5+1}}$, $\mathbf{\M E_{4+2}}$, $\mathbf{\M Ef_{6+1}}$ and $\mathbf{\M Ef_{5+2}}$ solvers
\begin{table}[t!]
\begin{center}
\setlength{\tabcolsep}{3.0pt}
\resizebox{1.0\columnwidth}{!}{
\begin{tabular}{|l|c|c|c|c|}
\hline
 Method & $\epsilon_{\textbf{R}}$ ($^\circ$) & $\epsilon_{\textbf{t}}$ ($^\circ$) & $\epsilon_{\textbf{p}}$ (\textit{m}) & $\epsilon_{\textbf{f}}$ (px) \\ \hline \rowcolor{mygray}
$\mathbf{s\M{H}5_3}$ & \best{0.21} / \phantom{1}\secondBest{0.45} & 1.31 / \phantom{1}4.51  & \secondBest{0.03} / 0.10 & -- \\ \hline
$\M{H}4$~\cite{hartley2003multiple} & 0.52 / \phantom{1}1.19  & 2.06 / \phantom{1}4.34 & 1.33 / 1.28 & -- \\ \hline
P3P + N & 0.39 / \phantom{1}0.91 & 1.84 / \phantom{1}6.43 &  \secondBest{0.03} / 0.10& -- \\ \hline
$\M{E}5$~\cite{stewenius2005minimal} & 0.49 / \phantom{1}1.18 & \best{1.21} / \phantom{1}\best{1.73} & \best{0.02} / \secondBest{0.03} & -- \\ \hline 
$\M{E}_{4+2}$~\cite{Wu_iccv15} & 0.49 / \phantom{1}1.19& 1.28 / \phantom{1}2.56 & \best{0.02} / \secondBest{0.03} & -- \\  \hline
$\M{E}_{4+2} + \text{mult. } \mathbf{\M{H}4}$ & 0.45 / \phantom{1}1.15  & \secondBest{1.24} / \phantom{1}2.16 & \secondBest{0.03} / 0.04 & -- \\ \hline   \rowcolor{mygray}
$\M{E}_{4+2} + \text{mult. } \mathbf{s\M{H}5_3}$ & {0.27} / \phantom{1}\best{0.32} & \secondBest{1.24} / \phantom{1}\secondBest{1.84} & \best{0.02} / \best{0.02} & -- \\ \hline \hline \rowcolor{mygray}
$\mathbf{s\M{H}5\textit{f}_3}$ & \secondBest{0.32} / \phantom{1}\best{1.09} & \secondBest{3.71} / \best{10.55} & \secondBest{0.05} / \secondBest{0.21} & \best{18.16} / \best{349.36}\\ \hline
P5P\textit{f} + N & \secondBest{0.25} / \secondBest{11.94} & \best{1.41} / \secondBest{10.80} & \best{0.02} / \best{0.05} & \phantom{1}\secondBest{40.66} / $>$\secondBest{10}$^{\secondBest{6}}$  \\ \hline 
\end{tabular}
}%
\end{center}
\vspace{-6pt}
\caption{Rotation, translation (degrees), position (meters) and focal length errors (pixels) on 
%23,190 
23k image pairs from KITTI. Best and second best results are shown in red and blue. 
Since most tested methods do not return the scale of the translation due to estimating the relative pose, we used the scale from the ground truth path to calculate the position error. Fig.~\ref{fig:kitti_experiments} shows the corresponding CDFs.}% %\todo{Swap median / mean}}
\label{tab:kitti_results}%
\end{table}

% \begin{figure}[t]
%     \centering
%         \includegraphics[trim={1mm 0mm 10mm 0mm},clip, width=0.30\textwidth]{LaTeX/images/cdf_focal_error.pdf} 
%     \caption{ The cum. distributions functions of the focal length errors on the $23,190$ image pairs from the KITTI dataset. A method being accurate is interpreted as a curve close to the top-left corner. }
%     \label{fig:kitti_experiments_focal}
%     \vspace{-6pt}
% \end{figure}

% \vspace{-7.5pt}
\PAR{Relative pose experiments.}
We use the 11 sequences of the KITTI benchmark~\cite{geiger2013vision} that are provided with the ground truth trajectories ($23,190$ image pairs).
In KITTI, the scenes are captured by two front-facing cameras mounted to a moving vehicle. 
We consider the camera pair as the generalized one and estimate the relative pose between this camera and the left image of the next frame.

As robust estimator, we use GC-RANSAC~\cite{barath2017graph} that applies two different solvers: (a) one for estimating the pose from a minimal sample and (b) one for fitting to a larger-than-minimal sample when polishing the model parameters on a set of inliers. 
We included the compared solvers in step (a). 
For step (b), we applied numerical optimization~\cite{ceres-solver}, minimizing the Sampson distance when estimating the essential matrix, and %the 
a re-projection error when estimating the homography.
Moreover, we test recovering the pose by combining the essential matrix from $\M{E}_{4+2}$ and multiple homographies~\cite{barath2019progressive} either from $\M{H}4$ or the proposed $\mathbf{s\M{H}5_3}$. 
The %outputted 
resulting pose is found by decomposing the essential matrix and homographies, and selecting the pose that has the largest support when thresholding the re-projection error. 

Tab.~\ref{tab:kitti_results} reports the median / mean rotation, position, 
% (degrees), position (meters), 
and focal length 
%(pixels) 
errors on the $23,190$ image pairs. 
Fig.~\ref{fig:kitti_experiments} shows the corresponding CDFs. % are shown in.
Since some of the tested methods, \eg, $\mathbf{\M{E}5}$, do not return the translation scale due to estimating the relative pose, we used the scale from the ground truth path to calculate the position error.
Even though the proposed $\mathbf{s\M{H}5_3}$ finds the most accurate rotation matrices, its translation and position errors are marginally higher than %that 
those of the essential matrix-based solvers.
Using $\mathbf{\M{E}_{4+2}}$ and multiple homographies from $\mathbf{s\M{H}5_3}$, however, leads to the most accurate poses. 
Amongst the solver estimating the focal length, the proposed $\mathbf{s\M{H}5f_3}$ solver is the most accurate one.
We did not include $\mathbf{\M Ef_{5+2}}$ and $\mathbf{\M Ef_{6}}$ since both %solvers 
fail when the camera undergoes purely forward motion.

%
%The cumulative distribution functions of the rotation, translation (both in degrees) and position (in meters) errors on the $23,190$ image pairs from the KITTI dataset are shown in Fig.~\ref{fig:kitti_experiments}.
%It can be seen that the proposed ${s\M{H}5_3}$ solver leads to the most accurate rotations. 
%The proposed
%${s\M{H}5f_3}$
%Hf3+2 
%solver is amongst the top-performing methods in terms of translation error. 
%
%The focal length errors for the proposed 
%${s\M{H}5f_3}$ and the 
%P5P$f$+N and ${\M{E}6f}$ methods are shown in Fig.~\ref{fig:kitti_experiments_focal}.
%As it is well-known~\cite{KukelovaP5pfr}, the ${\M{E}6f}$ solvers does not provide a reasonable focal length when the cameras undergo forward motion, which is often the case when having vehicle-mounted cameras.
%The proposed ${s\M{H}5f_3}$ solver estimates significantly more accurate focal lengths than the P5P$f$+N solver. 

\section{Conclusion}
\noindent
In this paper, 
we have considered the problem of estimating the  semi-generalized homography between a pinhole and a generalized camera. 
We have proposed efficient solvers handling both calibrated 
%pinhole  cameras 
and partially calibrated pinhole cameras with unknown focal length. 
Our solvers cover all possible minimal combinations of point correspondences between the pinhole and the generalized camera where it is possible to recover the scale. 
To the best of our knowledge, we are the first to solve this problem. 
% Through s
Synthetic 
%experiments 
and real experiments focusing on two real-world applications show that our solvers are practically relevant. %, %(localization and relative pose estimation for self-driving cars), 
% we have shown that our proposed solvers are of practical relevance. 
While they may not 
%be able to 
outperform more general existing solvers, which handle non-planar scenes, under all conditions, our results show that our solvers are preferable in certain conditions. 
Combining all these %types of 
solvers into a single hybrid RANSAC approach is thus an interesting direction for future work.

% \section{Acknowledgements} 
{
\small{
\PAR{Acknowledgements.} This paper was funded by the
OP VVV funded project CZ.02.1.01/0.0/0.0/16$\_$019/0000765 “Research
Center for Informatics”, the EU Horizon 2020 project RICAIP (%grant agreement 
No 857306) and the European Regional Development Fund under project IMPACT (No. CZ.02.1.01/0.0/0.0/15$\_$003/0000468). 
}
}

{\small
\bibliographystyle{ieee_fullname}
\bibliography{egbib,torsten}
}

\end{document}